\documentclass[sigconf]{acmart}

\usepackage{booktabs} 
\usepackage{epsfig}
\usepackage{graphicx}
\usepackage{amsmath}
\usepackage{amssymb}
\usepackage{bm}

\usepackage{algorithm}
\usepackage{algorithmic}
\usepackage{url}
\usepackage{subfigure}
\usepackage{bbding}
\usepackage{multirow}
\usepackage{enumitem}
\usepackage{array}
\setcopyright{rightsretained}

\copyrightyear{2018}
\acmYear{2018}
\setcopyright{acmcopyright}
\acmConference[MM '18]{2018 ACM Multimedia Conference}{Oct. 22--26, 2018}{Seoul, Republic of Korea}
\acmPrice{15.00}
\acmDOI{10.1145/3240508.3240512}
\acmISBN{978-1-4503-5665-7/18/10}

\begin{document}
\title{Visual Domain Adaptation with Manifold Embedded Distribution Alignment}
\titlenote{The first two authors contributed equally.}

\author{Jindong Wang, Wenjie Feng, Yiqiang Chen}
\authornote{J. Wang and Y. Chen are also affiliated with Beijing Key Lab. of Mobile Computing and Pervasive Devices. W. Feng is also with CAS Key Lab. of Network Data Science \& Technology. J. Wang and W. Feng are also affiliated with University of Chinese Academy of Sciences.}
\affiliation{%
  \institution{Institute of Computing Technology, CAS, Beijing, China}
}
\email{{wangjindong,yqchen}@ict.ac.cn}
\email{fengwenjie@software.ict.ac.cn}

\author{Han Yu}
\affiliation{%
	\institution{School of Computer Science and Engineering, NTU}
	\city{Singapore}
}

\email{han.yu@ntu.edu.sg}

\author{Meiyu Huang}
\affiliation{%
	\institution{Qian Xuesen Lab. of Space Technology, CAST}
	\city{Beijing, China}
}
\email{huangmeiyu@qxslab.cn}

\author{Philip S. Yu}
\authornote{P. Yu is also affiliated with Institute for Data Science, Tsinghua University, Beijing, China.}
\affiliation{%
	\institution{Department of Computer Science, UIC}
	\city{Chicago, USA}
}

\email{psyu@uic.edu}
%
%
%
%
%
%
%


\begin{abstract}
Visual domain adaptation aims to learn robust classifiers for the target domain by leveraging knowledge from a source domain. Existing methods either attempt to align the cross-domain distributions, or perform manifold subspace learning. However, there are two significant challenges: (1) \textit{degenerated feature transformation}, which means that distribution alignment is often performed in the original feature space, where feature distortions are hard to overcome. On the other hand, subspace learning is not sufficient to reduce the distribution divergence. (2) \textit{unevaluated distribution alignment}, which means that existing distribution alignment methods only align the marginal and conditional distributions with equal importance, while they fail to evaluate the different importance of these two distributions in real applications. In this paper, we propose a \textbf{Manifold Embedded Distribution Alignment~(MEDA)} approach to address these challenges. MEDA learns a domain-invariant classifier in Grassmann manifold with structural risk minimization, while performing dynamic distribution alignment to quantitatively account for the relative importance of marginal and conditional distributions. To the best of our knowledge, MEDA is the \textit{first} attempt to perform dynamic distribution alignment for manifold domain adaptation. Extensive experiments demonstrate that MEDA shows significant improvements in classification accuracy compared to state-of-the-art traditional and deep methods.
\end{abstract}

%
%
%


\maketitle

\section{Introduction}

The rapid growth of online media and content sharing applications has stimulated a great demand for automatic recognition and analysis for images and other multimedia data~\cite{ionescu2018datasets,chen2018zero}. Unfortunately, it is often expensive and time-consuming to acquire sufficient labeled data to train machine learning models. Thus, it is often necessary to leverage the abundant labeled samples in some existing domains to facilitate learning in a new target domain. Domain adaptation~\cite{pan2010survey,transferlearning} has been a promising approach to solve such cross-domain learning problems. 

Since the distributions of the source and target domains are different, the key to successful adaptation is to reduce the distribution divergence. To this end, existing work can be summarized into two main categories: (a)~\textit{instance reweighting}~\cite{dai2007boosting,xu2017unified}, which reuses samples from the source domain according to some weighting technique; and (b)~\textit{feature matching}, which either performs subspace learning by exploiting the subspace geometrical structure~\cite{fernando2013unsupervised,sun2016return,gong2012geodesic}, or distribution alignment to reduce the marginal or conditional distribution divergence between domains~\cite{zhang2017joint,long2013transfer}. Our focus is on feature matching methods. There are two significant challenges in existing methods, i.e. \textit{degenerated feature transformation} and \textit{unevaluated distribution alignment}. 

\textit{Degenerated feature transformation} means that both subspace learning and distribution alignment can only reduce, but not remove the distribution divergence~\cite{aljundi2015landmarks}. Specifically, subspace learning~\cite{gong2012geodesic,sun2016return,fernando2013unsupervised} conducts subspace transformation to obtain better feature representations. However, feature divergence is not eliminated after subspace transformation~\cite{long2014adaptation} since subspace learning only utilizes the subspace or manifold structure, but fails to perform feature alignment. On the other hand, distribution alignment~\cite{pan2011domain,long2013transfer,wang2017balanced} usually reduces the distribution distance in the original feature space, where features are often distorted~\cite{baktashmotlagh2013unsupervised} which makes it hard to reduce the divergence between domains. Therefore, it is critical to exploit both the advantages of subspace learning and distribution alignment to further facilitate domain adaptation.

\begin{figure}[t!]
	\centering
	\vspace{-.1in}
	\subfigure[Source]{
		\includegraphics[scale=0.35]{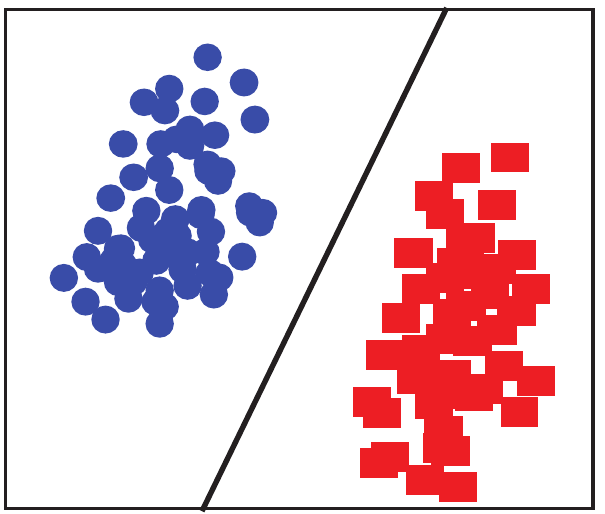}
		\label{fig-sub-source-a}}
	~\vline
	~
	\subfigure[Target: Type I]{
		\includegraphics[scale=0.35]{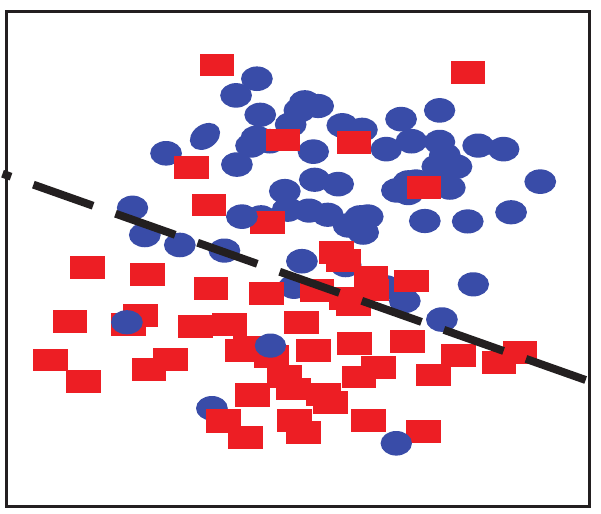}
		\label{fig-sub-source-b}}
	~
	\subfigure[Target: Type II]{
		\includegraphics[scale=0.35]{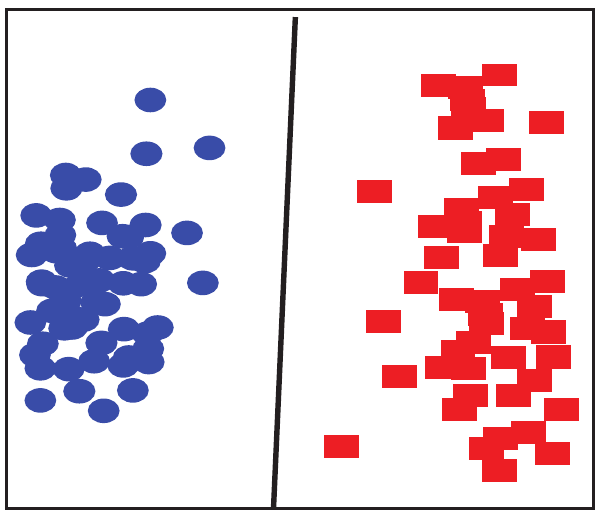}
		\label{fig-sub-source-c}}
	\vspace{-.1in}
	\caption{Examples of two different target domains w.r.t. the same source domain during distribution adaption.}
	\label{fig-bda}
	\vspace{-.2in}
\end{figure}

\textit{Unevaluated distribution alignment} means that existing work~\cite{long2013transfer,hou2016unsupervised,zhang2017joint,tahmoresnezhad2016visual} only attempted to align the marginal and conditional distributions with \textit{equal} weights. But they failed to evaluate the relative importance of these two distributions. For example, when two domains are very dissimilar~(Figure~\ref{fig-sub-source-a} $\rightarrow$ \ref{fig-sub-source-b}), the marginal distribution is more important to align. When the marginal distributions are close~(Figure~\ref{fig-sub-source-a} $\rightarrow$ \ref{fig-sub-source-c}), the conditional distribution should be given more weight. However, there is no alignment method which can \textit{quantitatively} account for the importance of these two distributions in conjunction.

As far as we know, there has been no previous work that tackle these two challenges together. In this paper, we propose a novel \textbf{Manifold Embedded Distribution Alignment~(MEDA)} method to address the challenges of both degenerated feature transformation and unevaluated distribution alignment. MEDA learns a domain-invariant classifier in Grassmann manifold with structural risk minimization, while performing dynamic distribution alignment by considering the different importance of marginal and conditional distributions. We also provide a feasible solution to quantitatively evaluate the importance of distributions. To the best of our knowledge, MEDA is the \textit{first} attempt to reveal the relative importance of marginal and conditional distributions in domain adaptation.

This work makes the following contributions:

1) We propose the MEDA approach for domain adaptation. MEDA is capable of addressing both the challenges of degenerated feature transformation and unevaluated distribution alignment.

2) We provide the \textit{first} quantitative evaluation of the relative importance of marginal and conditional distributions in domain adaptation. This is significantly useful in future research on transfer learning.

3) Extensive experiments on 7 real-world image datasets demonstrate that compared to several state-of-the-art traditional and deep methods, MEDA achieves a significant improvement of $\bm{3.5}$\textbf{\%} in average classification accuracy.

\section{Related Work}
\label{sec-related}

MEDA substantially distinguishes from existing feature matching domain adaptation methods in several aspects:

\textit{Subspace learning.} Subspace Alignment~(SA)~\cite{fernando2013unsupervised} aligned the base vectors of both domains, but failed to adapt feature distributions. Subspace distribution alignment (SDA)~\cite{sun2015subspace} extended SA by adding the subspace variance adaptation. However, SDA did not consider the local property of subspaces and ignored conditional distribution alignment. CORAL~\cite{sun2016return} aligned subspaces in second-order statistics, but it did not consider the distribution alignment. Scatter component analysis~(SCA)~\cite{ghifary2017scatter} converted the samples into a set of subspaces ~(i.e. scatters) and then minimized the divergence between them. GFK~\cite{gong2012geodesic} extended the idea of sampled points in manifold~\cite{gopalan2011domain} and proposed to learn the geodesic flow kernel between domains. The work of \cite{baktashmotlagh2014domain} used a Hellinger distance to approximate the geodesic distance in Riemann space. \cite{baktashmotlagh2013unsupervised} proposed to use Grassmann for domain adaptation, but they ignored the conditional distribution alignment. Different from these approaches, MEDA can learn a domain-invariant classifier in the manifold and align both marginal and conditional distributions.

\textit{Distribution alignment.} MEDA substantially differs from existing work that only align marginal or conditional distribution~\cite{pan2011domain}. Joint distribution adaptation~(JDA)~\cite{long2013transfer} proposed to match both distributions with equal weights. Others extended JDA by adding regularization~\cite{long2014adaptation}, sparse representation~\cite{xu2016discriminative}, structural consistency~\cite{hou2016unsupervised}, domain invariant clustering~\cite{tahmoresnezhad2016visual}, and label propagation~\cite{zhang2017joint}. The main differences between MEDA and these methods are: 1) These work treats the two distributions equally. However, when there is a greater discrepancy between both distributions, they cannot evaluate their relative importance and thus lead to undermined performance. Our work is capable of evaluating the quantitative importance of each distribution via considering their different effects. 2) These methods are designed only for the original space, where feature distortion will hinder the performance. MEDA can align the distributions in the manifold to overcome the feature distortions.

\textit{Domain-invariant classifier learning.} The recent work of ARTL~\cite{long2014adaptation}, DIP~\cite{baktashmotlagh2013unsupervised,baktashmotlagh2016distribution}, and DMM~\cite{cao2018unsupervised} also aimed to build a domain-invariant classifier. However, ARTL and DMM can be undermined by feature distortion in original space, and they failed to leverage the different importance of distributions. DIP mainly focused on feature transformation and only aligned marginal distributions. MEDA is able to avoid the feature distortion and quantitatively evaluate the importance of marginal and conditional distribution alignment.

\section{Manifold Embedded distribution alignment}
\label{sec-method}

In this section, we present the Manifold Embedded distribution alignment~(MEDA) approach in detail.

\subsection{Problem Definition}
Given a labeled source domain $\mathcal{D}_s=\{\mathbf{x}_{s_i},y_{s_i}\}^n_{i=1}$ and an unlabeled target domain $\mathcal{D}_t=\{\mathbf{x}_{t_j}\}^{n+m}_{j=n+1}$, assume the feature space $\mathcal{X}_s = \mathcal{X}_t$, label space $\mathcal{Y}_s = \mathcal{Y}_t$, but marginal probability $P_s(\mathbf{x}_s) \ne P_t(\mathbf{x}_t)$ with conditional probability $Q_s(y_s|\mathbf{x}_s) \ne Q_t(y_t|\mathbf{x}_t)$. The goal of domain adaptation is to learn a classifier $f:\mathbf{x}_t \mapsto \mathbf{y}_t$ to predict the labels $\mathbf{y}_t \in \mathcal{Y}_t$ for the target domain $\mathcal{D}_t$ using labeled source domain $\mathcal{D}_s$.

According to the structural risk minimization~(SRM)~\cite{vapnik1998statistical}, $f = \mathop{\arg\min}_{f \in \mathcal{H}_{K}} \ell(f(\mathbf{x}),\mathbf{y}) + R(f)$, where the first term indicates the loss on data samples, the second term denotes the regularization term, and $\mathcal{H}_{K}$ is the Hilbert space induced by kernel function $K(\cdot,\cdot)$. Since there is no labels on $\mathcal{D}_t$, we can only perform SRM on $\mathcal{D}_s$. Moreover, due to the different distributions between $\mathcal{D}_s$ and $\mathcal{D}_t$, it is necessary to add other constraints to maximize the distribution consistency while learning $f$.

\begin{figure}[t]
	\centering
	\includegraphics[scale=0.66]{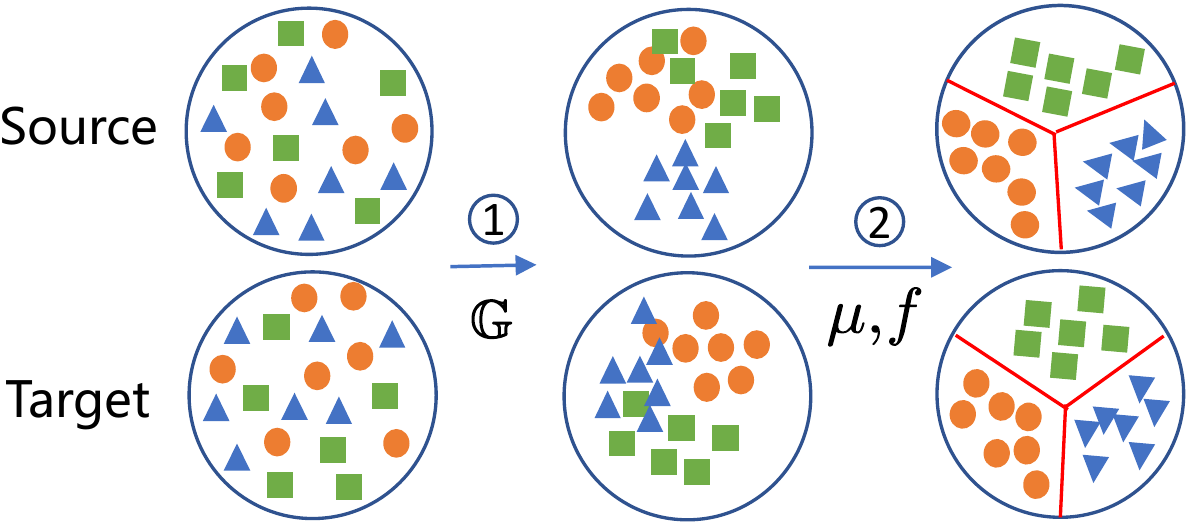}
	\caption{The main idea of MEDA. \textcircled{1} Features in the original space are transformed into manifold space by learning the manifold kernel $\mathbb{G}$. \textcircled{2} Dynamic distribution alignment (by learning $\mu$) with SRM is performed in manifold to learn the final domain-invariant classifier $f$.}
	\label{fig-main}
\end{figure}

\subsection{Main Idea}

MEDA consists of two fundamental steps. Firstly, MEDA performs \textit{manifold feature learning} to address the challenge of degenerated feature transformation. Secondly, MEDA performs \textit{dynamic distribution alignment} to quantitatively account for the relative importance of marginal and conditional distributions to address the challenge of unevaluated distribution alignment. Eventually, a domain-invariant classifier $f$ can be learned by summarizing these two steps with the principle of SRM. Figure~\ref{fig-main} presents the main idea of the proposed MEDA approach. 

Formally, if we denote $g(\cdot)$ the manifold feature learning functional, then $f$ can be represented as

\begin{equation}
\label{equ-f-orig}
\begin{split}
f = \mathop{\arg\min}_{f \in \sum_{i=1}^{n} \mathcal{H}_{K}} &\ell(f(g(\mathbf{x}_i)),y_i) + \eta ||f||^2_K \\
&+ \lambda \overline{D_f}(\mathcal{D}_s,\mathcal{D}_t) + \rho R_f(\mathcal{D}_s,\mathcal{D}_t)
\end{split}
\end{equation}
where $||f||^2_K$ is the squared norm of $f$. The term $\overline{D_f}(\cdot,\cdot)$ represents the proposed dynamic distribution alignment. Additionally, we introduce $R_f(\cdot,\cdot)$ as a Laplacian regularization to further exploit the similar geometrical property of nearest points in manifold $\mathbb{G}$~\cite{belkin2006manifold}. $\eta,\lambda$, and $\rho$ are regularization parameters accordingly.

The overall learning process of MEDA is in Algorithm~\ref{algo-meda}. In next sections, we first introduce manifold feature learning (learn $g(\cdot)$). Then, we present the dynamic distribution alignment (learn $\overline{D_f}(\cdot,\cdot)$). Eventually, we articulate the learning of $f$.

\subsection{Manifold Feature Learning}

Manifold feature learning serves as the preprocessing step to eliminate the threat of \textit{degenerated feature transformation}. MEDA learns $g(\cdot)$ in the \textit{Grassmann} manifold $\mathbb{G}(d)$~\cite{hamm2008grassmann} since features in the manifold have some geometrical structures~\cite{belkin2006manifold,hamm2008grassmann} that can avoid distortion in the original space. And $\mathbb{G}$ can facilitate classifier learning by treating the original $d$-dimensional subspace (i.e. feature vector) as its basic element~\cite{baktashmotlagh2014domain}. Additionally, feature transformation and distribution alignment often have efficient numerical forms and can thus facilitate domain adaptation on $\mathbb{G}(d)$~\cite{hamm2008grassmann}. There are several approaches to transform the features into $\mathbb{G}$ \cite{gopalan2011domain,baktashmotlagh2014domain}, among which we embed Geodesic Flow Kernel (GFK)~\cite{gong2012geodesic} to learn $g(\cdot)$ for its computational efficiency. We only introduce the main idea of GFK and the details can be found in its original paper.

When learning manifold features, MEDA tries to model the domains with $d$-dimensional subspaces and then embed them into $\mathbb{G}$. Let $\mathcal{S}_s$ and $\mathcal{S}_t$ denote the PCA subspaces for the source and target domain, respectively. $\mathbb{G}$ can thus be regarded as a collection of all $d$-dimensional subspaces. Each original subspace can be seen as a point in $\mathbb{G}$. Therefore, the geodesic flow $\{\Phi(t):0 \leq t \leq 1\}$ between two points can draw a path for the two subspaces. If we let $\mathcal{S}_s=\Phi(0)$ and $\mathcal{S}_t=\Phi(1)$, then finding a geodesic flow from $\Phi(0)$ to $\Phi(1)$ equals to transforming the original features into an infinite-dimensional feature space, which eventually eliminates the domain shift. This kind of approach can be seen as an incremental way of `walking' from $\Phi(0)$ to $\Phi(1)$. Specifically, the new features can be represented as $\mathbf{z}=g(\mathbf{x}) = \Phi(t)^T \mathbf{x}$. From \cite{gong2012geodesic}, the inner product of transformed features $\mathbf{z}_i$ and $\mathbf{z}_j$ gives rise to a positive semidefinite geodesic flow kernel:
\begin{equation}
	\label{equ-gfk}
	\langle\mathbf{z}_i,\mathbf{z}_j\rangle= \int_{0}^{1} (\Phi(t)^T \mathbf{x}_i)^T (\Phi(t)^T \mathbf{x}_j) \, dt = \mathbf{x}^T_i \mathbf{G} \mathbf{x}_j
\end{equation}

Thus, the feature in original space can be transformed into Grassmann manifold with $\mathbf{z}=g(\mathbf{x}) = \sqrt{\mathbf{G}}\mathbf{x}$. $\mathbf{G}$ can be computed efficiently by singular value decomposition~\cite{gong2012geodesic}. Note that $\sqrt{\mathbf{G}}$ is only an expression form and cannot be computed directly, while its square root is calculated by Denman-Beavers algorithm~\cite{denman1976matrix}.

\label{sec-da}
\subsection{Dynamic Distribution Alignment}

The purpose of dynamic distribution alignment is to \textit{quantitatively} evaluate the importance of aligning marginal~($P$) and conditional~($Q$) distributions in domain adaptation.
Existing methods~\cite{long2013transfer,zhang2017joint} failed in this evaluation by only assuming that both distributions are equally important. However, this assumption may not be realistic for real applications. For instance, when transferring from Figure \ref{fig-sub-source-a} to \ref{fig-sub-source-b}, there is a large difference between datasets. Therefore, the divergence between $P_s$ and $P_t$ is more dominant. In contrast, from Figure~\ref{fig-sub-source-a} to~\ref{fig-sub-source-c}, the datasets are similar. Therefore, the distribution divergence in each class ($Q_s$ and $Q_t$) is more dominant. 

\textbf{The adaptive factor:}

In view of this phenomenon, we introduce an \textit{adaptive factor} to dynamically leverage the importance of these two distributions. Formally, the dynamic distribution alignment $\overline{D_f}$ is defined as
\begin{equation}
	\label{equ-bda}
	\overline{D_{f}}(\mathcal{D}_s,\mathcal{D}_t) = (1- \mu)D_{f}(P_s,P_t) + \mu \sum_{c=1}^{C} D^{(c)}_{f}(Q_s,Q_t)
\end{equation}
where $\mu \in [0,1]$ is the adaptive factor and $c \in \{1,\cdots,C\}$ is the class indicator. $D_f(P_s,P_t)$ denotes the marginal distribution alignment, and $D^{(c)}_f(Q_s,Q_t)$ denotes the conditional distribution alignment for class $c$.

When $\mu \rightarrow 0$, it means that the distribution distance between the source and the target domains is large. Thus, marginal distribution alignment is more important~(Figure~\ref{fig-sub-source-a} $\rightarrow$ \ref{fig-sub-source-b}). When $\mu \rightarrow 1$, it means that feature distribution between domains is relatively small, so the distribution of each class is dominant. Thus, the conditional distribution alignment is more important (Figure~\ref{fig-sub-source-a} $\rightarrow$ \ref{fig-sub-source-c}). When $\mu=0.5$, both distributions are treated equally as in existing methods~\cite{long2013transfer,zhang2017joint}. Hence, the existing methods can be regarded as the special cases of MEDA. By learning the optimal adaptive factor $\mu_{opt}$ (which we will discuss later), MEDA can be applied to different domain adaptation problems.

We use the \textit{maximum mean discrepancy} (MMD)~\cite{ben2007analysis} to empirically calculate the distribution divergence between domains. As a nonparametric measurement, MMD has been widely applied in many existing methods~\cite{zhang2017joint,ghifary2017scatter,pan2011domain}, and its theoretical effectiveness has been verified in~\cite{gretton2012kernel}. The MMD distance between distributions $p$ and $q$ is defined as $d^2(p,q)=(\mathbb{E}_p[\phi(\mathbf{z}_s)] - \mathbb{E}_q[\phi(\mathbf{z}_t)])^2_{\mathcal{H}_{K}}$ where $\mathcal{H}_{K}$ is the reproducing kernel Hilbert space (RKHS) induced by feature map $\phi(\cdot)$. Here, $\mathbb{E}[\cdot]$ denotes the mean of the embedded samples. In order to compute an MMD associated with $f$, we adopt \textit{projected MMD}~\cite{quanz2009large} and compute the marginal distribution alignment as $D_f(P_s,P_t)=\Vert\mathbb{E}[f(\mathbf{z}_s)] - \mathbb{E}[f(\mathbf{z}_t)]\Vert^2_{\mathcal{H}_{K}}$. Similarly, the conditional distribution alignment is $D^{(c)}_f(Q_s,Q_t)=\Vert \mathbb{E}[f(\mathbf{z}^{(c)}_s)] - \mathbb{E}[f(\mathbf{z}^{(c)}_t)]\Vert^2_{\mathcal{H}_{K}}$. Then, dynamic distribution alignment can be expressed as
\begin{equation}
	\label{equ-da1}
	\begin{split}
		\overline{D_{f}}(\mathcal{D}_s,\mathcal{D}_t) = &(1 - \mu) \Vert\mathbb{E}[f(\mathbf{z}_s)) - \mathbb{E}[f(\mathbf{z}_t)]\Vert^2_{\mathcal{H}_{K}} \\
		&+ \mu \sum_{c=1}^{C}\Vert \mathbb{E}[f(\mathbf{z}^{(c)}_s)] - \mathbb{E}[f(\mathbf{z}^{(c)}_t)]\Vert^2_{\mathcal{H}_{K}}
	\end{split}
\end{equation}

Note that since $\mathcal{D}_t$ has no labels, it is not feasible to evaluate the conditional distribution $Q_t=Q_t(y_t|\mathbf{z}_t)$. Instead, we follow the idea in~\cite{wang2017balanced} and use the class conditional distribution $Q_t(\mathbf{z}_t|y_t)$ to approximate $Q_t$. In order to evaluate $Q_t(\mathbf{z}_t|y_t)$, we apply prediction to $\mathcal{D}_t$ using a base classifier trained on $\mathcal{D}_s$ to obtain soft labels for $\mathcal{D}_t$. The soft labels may be less reliable, so we \textit{iteratively} refine the prediction. Note that we \textit{only} use the base classifier in the first iteration. After that, MEDA can \textit{automatically} refine the labels for $\mathcal{D}_t$ using results from previous iteration.

\textbf{The quantitative evaluation of the adaptive factor} $\bm{\mu}$:

We can treat $\mu$ as a parameter and tune its value by cross-validation techniques. However, there is no labels for the target domain in unsupervised domain adaptation problems. It is extremely hard to calculate the value of $\mu$. In this work, we made the \textit{first} attempt towards calculating $\mu$ (i.e. $\hat{\mu}$) by exploiting the global and local structure of domains. We adopted the $\mathcal{A}$-distance~\cite{ben2007analysis} as the basic measurement. The $\mathcal{A}$-distance is defined as the error of building a linear classifier to discriminate two domains (i.e. a binary classification). Formally, we denote $\epsilon(h)$ the error of a linear classifier $h$ discriminating the two domains $\mathcal{D}_s$ and $\mathcal{D}_t$. Then, the $\mathcal{A}$-distance can be defined as
\begin{equation}
	d_A(\mathcal{D}_s,\mathcal{D}_t) = 2(1 - 2 \epsilon(h))
\end{equation}

We can directly compute the marginal $\mathcal{A}$-distance using above equation, which is denoted as $d_M$. For the $\mathcal{A}$-distance between conditional distributions, we denote $d_c$ as the $\mathcal{A}$-distance for the $c$th class. It can be calculated as $d_c = d_A(\mathcal{D}^{(c)}_s,\mathcal{D}^{(c)}_t)$, where $\mathcal{D}^{(c)}_s$ and $\mathcal{D}^{(c)}_t$ denote samples from class $c$ in $\mathcal{D}_s$ and $\mathcal{D}_t$, respectively. Eventually, $\mu$ can be estimated as

\begin{equation}
	\label{eq-mu}
	\hat{\mu} \approx 1 - \frac{d_M}{d_M + \sum_{c=1}^{C} d_c}
\end{equation}

This estimation has to be conducted at every iteration of the dynamic distribution adaptation, since the feature distribution may vary after evaluating the conditional distribution each time. 
To be noticed, this is the \textit{first} solution to quantitatively estimate the relative importance of each distribution. In fact, this estimation can be of significant help in future research on transfer learning and domain adaptation.

\subsection{Learning Classifier $f$}

After manifold feature learning and dynamic distribution alignment, $f$ can be learned by summarizing SRM over $\mathcal{D}_s$ and distribution alignment. Adopting the square loss $l_2$, $f$ can be represented as
\begin{equation}
	\label{equ-f}
	\begin{split}
		f = \mathop{\arg\min}_{f \in \mathcal{H}_{K}} \sum_{i=1}^{n} (y_i &- f(\mathbf{z}_i))^2 + \eta ||f||^2_K \\
		&+ \lambda \overline{D_f}(\mathcal{D}_s,\mathcal{D}_t) + \rho R_f(\mathcal{D}_s,\mathcal{D}_t)
	\end{split}
\end{equation}

In order to perform efficient learning, we now reformulate each term in detail.

\textbf{SRM on the Source Domain:} 
Using the representer theorem~\cite{belkin2006manifold}, $f$ admits the expansion 
\begin{equation}
	\label{equ-repr}
	f(\mathbf{z})=\sum_{i=1}^{n+m} \beta_i K(\mathbf{z}_i,\mathbf{z})
\end{equation}
where $\bm{\beta}=(\beta_1,\beta_2,\cdots)^T \in \mathbb{R}^{(n + m) \times 1}$ is the coefficients vector and $K$ is a kernel. Then, SRM on $\mathcal{D}_s$ can be
\begin{equation}
	\label{equ-risk2}
	\begin{split}
		&\sum_{i=1}^{n} (y_i - f(\mathbf{z}_i))^2 + \eta ||f||^2_K \\
		&= \sum_{i=1}^{n+m} \mathbf{A}_{ii}(y_i - f(\mathbf{z}_i))^2 + \eta ||f||^2_K\\
		&= ||(\mathbf{Y} - \bm{\beta}^T \mathbf{K}) \mathbf{A}||^2_{F} + \eta \mathrm{tr}(\bm{\beta}^T \mathbf{K} \bm{\beta})
	\end{split}	
\end{equation}
where $||\cdot||_F$ is the Frobenious norm. $\mathbf{K} \in \mathbb{R}^{(n+m) \times (n+m)}$ is the kernel matrix with $\mathbf{K}_{ij}=K(\mathbf{z}_i,\mathbf{z}_j)$, and $\mathbf{A} \in \mathbb{R}^{(n+m) \times (n+m)}$ is a diagonal domain indicator matrix with $\mathbf{A}_{ii}=1$ if $i \in \mathcal{D}_s$, otherwise $\mathbf{A}_{ii}=0$. $\mathbf{Y}=[y_1,\cdots,y_{n+m}]$ is the label matrix from source and the target domains. $\mathrm{tr(\cdot)}$ denotes the trace operation. Although the labels for $\mathcal{D}_t$ are unavailable, they can be filtered out by the indicator matrix $\mathbf{A}$.

\textbf{Dynamic distribution alignment: }
Using the representer theorem and kernel tricks, dynamic distribution alignment in equation~(\ref{equ-da1}) becomes
\begin{equation}
	\label{equ-da}
	\overline{D_{f}}(\mathcal{D}_s,\mathcal{D}_t)=\mathrm{tr} \left(\bm{\beta}^T \mathbf{K} \mathbf{M} \mathbf{K} \bm{\beta} \right)
\end{equation}
where $\mathbf{M}=(1-\mu)\mathbf{M}_0 + \mu \sum_{c=1}^{C} \mathbf{M}_c$ is the MMD matrix with its element calculated by
\begin{equation}
	\label{equ-mo}
	(\mathbf{M}_0)_{ij}=\begin{cases}
		\frac{1}{n^2},  & \mathbf{z}_i,\mathbf{z}_j \in \mathcal{D}_s\\ 
		\frac{1}{m^2}, & \mathbf{z}_i,\mathbf{z}_j \in \mathcal{D}_t\\ 
		-\frac{1}{mn}, & \text{otherwise} 
	\end{cases}
\end{equation}
\begin{equation}
	\label{equ-mc}
	(\mathbf{M}_c)_{ij}=\begin{cases}
		\frac{1}{n^2_c},  & \mathbf{z}_i,\mathbf{z}_j \in \mathcal{D}^{(c)}_s\\ 
		\frac{1}{m^2_c}, & \mathbf{z}_i,\mathbf{z}_j \in \mathcal{D}^{(c)}_t\\ 
		-\frac{1}{m_c n_c}, & \begin{cases}
			\mathbf{z}_i \in \mathcal{D}^{(c)}_s ,\mathbf{z}_j \in \mathcal{D}^{(c)}_t \\ 
			\mathbf{z}_i \in \mathcal{D}^{(c)}_t ,\mathbf{z}_j \in \mathcal{D}^{(c)}_s
		\end{cases}\\
		0, & \text{otherwise}
	\end{cases}
\end{equation}
where $n_c=|\mathcal{D}^{(c)}_s|$ and $m_c=|\mathcal{D}^{(c)}_t|$.

\textbf{Laplacian Regularization: } 
Additionally, we add a Laplacian regularization term to further exploit the similar geometrical property of nearest points in manifold $\mathbb{G}$~\cite{belkin2006manifold}. We denote the pair-wise affinity matrix as
\begin{equation}
	\mathbf{W}_{ij} = \begin{cases}
		\mathrm{sim}(\mathbf{z}_i,\mathbf{z}_j), & \mathbf{z}_i \in \mathcal{N}_p(\mathbf{z}_j) \text{ or } \mathbf{z}_j \in \mathcal{N}_p(\mathbf{z}_i) \\
		0, & \text{otherwise}
	\end{cases}
\end{equation}
where $\mathrm{sim}(\cdot,\cdot)$ is a similarity function~(such as cosine distance) to measure the distance between two points. $\mathcal{N}_p(\mathbf{z}_i)$ denotes the set of $p$-nearest neighbors to point $\mathbf{z}_i$. $p$ is a free parameter and must be set in the method. By introducing Laplacian matrix $\mathbf{L}=\mathbf{D} - \mathbf{W}$ with diagonal matrix $\mathbf{D}_{ii}=\sum_{j=1}^{n+m} \mathbf{W}_{ij}$, the final regularization can be expressed by
\begin{equation}
	\label{equ-lap}
	\begin{split}
		R_f(\mathcal{D}_s,\mathcal{D}_t)&=\sum_{i,j=1}^{n+m} \mathbf{W}_{ij} (f(\mathbf{z}_i)-f(\mathbf{z}_j))^2\\
		&=\sum_{i,j=1}^{n+m} f(\mathbf{z}_i) \mathbf{L}_{ij} f(\mathbf{z}_j)\\
		&=\mathrm{tr} \left(\bm{\beta}^T \mathbf{K} \mathbf{L} \mathbf{K} \bm{\beta}\right)
	\end{split}
\end{equation}

\textbf{Overall Reformulation:} Substituting with equations~(\ref{equ-risk2}), (\ref{equ-da}) and (\ref{equ-lap}), $f$ in equation (\ref{equ-f}) can be reformulated as
\begin{equation}
	\label{equ-final}
	\begin{split}
		f=\mathop{\arg\min}_{f \in \mathcal{H}_{K}}||(\mathbf{Y} &- \bm{\beta}^T \mathbf{K}) \mathbf{A}||^2_{F} + \eta \, \mathrm{tr}(\bm{\beta}^T \mathbf{K} \bm{\beta})\\
		&+ \mathrm{tr}\left(\bm{\beta}^T \mathbf{K}(\lambda \mathbf{M} + \rho \mathbf{L}) \mathbf{K} \bm{\beta} \right)
	\end{split}
\end{equation}
Setting derivative $\partial f/ \partial \bm{\beta}=0$, we obtain the solution 
\begin{equation}
	\label{equ-solution}
	\bm{\beta}^\star= ((\mathbf{A} + \lambda \mathbf{M} + \rho \mathbf{L}) \mathbf{K} + \eta \mathbf{I})^{-1} \mathbf{A} \mathbf{Y}^T
\end{equation}

MEDA has a nice property: it can learn the cross-domain function directly \textit{without} the need of explicit classifier training. This makes it significantly different from most existing work such as JGSA~\cite{zhang2017joint} and CORAL~\cite{sun2016return} that further needs to learn a certain classifier.

\begin{algorithm}[t!] 
	\caption{Manifold Embedded Distribution Alignment}  
	\label{algo-meda}  
	\renewcommand{\algorithmicrequire}{\textbf{Input:}} 
	\renewcommand{\algorithmicensure}{\textbf{Output:}}
	\begin{algorithmic}[1]  
		\REQUIRE 
		Data matrix $\mathbf{X}=[\mathbf{X}_s,\mathbf{X}_t]$, source domain labels $\mathbf{y}_s$, manifold subspace dimension $d$, regularization parameters $\lambda,\eta,\rho$, and \#neighbor $p$.\\
		\ENSURE 
		Classifier $f$.\\
		\STATE Learn manifold feature transformation kernel~$\mathbf{G}$ via equation~(\ref{equ-gfk}), and get manifold feature $\mathbf{Z}=\sqrt{\mathbf{G}} \mathbf{X}$.
		\STATE Train a base classifier using $\mathcal{D}_s$, then apply prediction on $\mathcal{D}_t$ to get its soft labels $\hat{y}_t$. 
		\STATE Construct kernel $\mathbf{K}$ using transformed features $\mathbf{Z}_s=\mathbf{Z}_{1:n,:}$ and $\mathbf{Z}_t=\mathbf{Z}_{n+1:n+m,:}$.
		\REPEAT 
		\STATE Calculate the adaptive factor $\hat{\mu}$ using equation~(\ref{eq-mu}).
		 and compute $\mathbf{M}_0$ and $\mathbf{M}_c$ by equations~(\ref{equ-mo}) and (\ref{equ-mc}).
		\STATE Compute $\bm{\beta}^\star$ by solving equation~(\ref{equ-solution}) and obtain $f$ via the representer theorem in equation~(\ref{equ-repr}).
		\STATE Update the soft labels of $\mathcal{D}_t$: $\hat{y}_t=f(\mathbf{Z}_t)$.
		\UNTIL{Convergence}  
		\RETURN Classifier $f$.  
	\end{algorithmic}  
\end{algorithm}

\section{Experiments and Evaluations}
\label{sec-exp}
In this section, we evaluate the performance of MEDA through extensive experiments on large-scale public datasets. The source code for MEDA is available at \url{http://transferlearning.xyz/}.

\subsection{Data Preparation}
\label{sec-exp-data}

We adopted seven publicly image datasets: Office+Caltech10, USPS + MNIST, ImageNet + VOC2007, and Office-31. These datasets are popular for benchmarking domain adaptation algorithms and have been widely adopted in most existing work such as~\cite{gong2012geodesic,long2014adaptation,zhang2017joint,zhuo2017deep}. Table~\ref{tb-dataset} lists the statistics of the seven datasets.

\textbf{Office-31}~\cite{saenko2010adapting} consists of three real-world object domains: \textbf{Amazon} (A), \textbf{Webcam} (W) and \textbf{DSLR} (D). It has 4,652 images with 31 categories. \textbf{Caltech-256} (C) contains 30,607 images and 256 categories. Since the objects in Office and Caltech follow different distributions, domain adaptation can help to perform cross-domain recognition. There are 10 common classes in the two datasets. For our experiments, we adopted the \textbf{Office+Caltech10} datasets from~\cite{gong2012geodesic} which contains 12 tasks: A $\rightarrow$ D, A $\rightarrow$ C,..., C $\rightarrow$ W. In the rest of the paper, we use $A \rightarrow B$ to denote the knowledge transfer from source domain \textit{A} to the target domain \textit{B}.

\textbf{USPS} (U) and \textbf{MNIST} (M) are standard digit recognition datasets containing handwritten digits from 0-9. Since the same digits across two datasets follow different distributions, it is necessary to perform domain adaptation. USPS consists of 7,291 training images and 2,007 test images of size 16 $\times$ 16. MNIST consists of 60,000 training images and 10,000 test images of size 28 $\times$ 28. We construct two tasks: U $\rightarrow$ M and M $\rightarrow$ U. 

\textbf{ImageNet} (I) and \textbf{VOC2007} (V) are large standard image recognition datasets. Each dataset can be treated as one domain. The images from the same classes of two domains follow different distributions. In our experiments, we adopt the sub-datasets presented in \cite{fang2013unbiased} to construct cross-domain tasks. Five common classes are extracted from both datasets: \textit{bird, cat, chair, dog}, and \textit{person}. Eventually, we have two tasks: I $\rightarrow$ V and V $\rightarrow$ I.
 
\begin{table}[t!]
	\centering
	\vspace{-.1in}
	\caption{Statistics of the seven benchmark datasets.}
	\vspace{-.15in}
	\label{tb-dataset}
	\resizebox{0.45\textwidth}{!}{
		\begin{tabular}{|c|c|c|c|c|}
			\hline
			\textbf{Dataset} & \textbf{\#Sample} & \textbf{\#Feature} & \textbf{\#Class} & \textbf{Domain} \\ \hline \hline
			Office-10 & 1,410 & 800 (4,096) & 10 & A, W, D \\ \hline
			Caltech-10 & 1,123 & 800 (4,096) & 10 & C \\ \hline
			Office-31 & 4,652 & 4,096 & 31 & A, W, D \\ \hline
			USPS & 1,800 & 256 & 10 & USPS (U) \\ \hline
			MNIST & 2,000 & 256 & 10 & MNIST (M) \\ \hline		
			ImageNet & 7,341 & 4,096 & 5 & ImageNet (I) \\ \hline
			VOC2007 & 3,376 & 4,096 & 5 & VOC (V) \\ \hline
		\end{tabular}
	}
\vspace{-.15in}
\end{table}

\begin{table*}[ht]
	\centering
	\vspace{-.1in}
	\caption{Accuracy (\%) on Office+Caltech10 datasets using SURF features.}
	\vspace{-.15in}
	\label{tb-surf}
	\resizebox{0.7\textwidth}{!}{
		\begin{tabular}{|c|c|c|c|c|c|c|c|c|c|c|c|c|}
			\hline
			Task & 1NN & SVM & PCA & TCA & GFK & JDA & TJM & CORAL & SCA & ARTL & JGSA & MEDA \\ \hline \hline
			C $\rightarrow$ A & 23.7 & 53.1 & 39.5 & 45.6 & 46.0 & 43.1 & 46.8 & 52.1 & 45.6 & 44.1 & 51.5 & \textbf{56.5} \\ \hline
			C $\rightarrow$ W & 25.8 & 41.7 & 34.6 & 39.3 & 37.0 & 39.3 & 39.0 & 46.4 & 40.0 & 31.5 & 45.4 & \textbf{53.9} \\ \hline
			C $\rightarrow$ D & 25.5 & 47.8 & 44.6 & 45.9 & 40.8 & 49.0 & 44.6 & 45.9 & 47.1 & 39.5 & 45.9 & \textbf{50.3} \\ \hline
			A $\rightarrow$ C & 26.0 & 41.7 & 39.0 & 42.0 & 40.7 & 40.9 & 39.5 & 45.1 & 39.7 & 36.1 & 41.5 & \textbf{43.9} \\ \hline
			A $\rightarrow$ W & 29.8 & 31.9 & 35.9 & 40.0 & 37.0 & 38.0 & 42.0 & 44.4 & 34.9 & 33.6 & 45.8 & \textbf{53.2} \\ \hline
			A $\rightarrow$ D & 25.5 & 44.6 & 33.8 & 35.7 & 40.1 & 42.0 & 45.2 & 39.5 & 39.5 & 36.9 & 47.1 & \textbf{45.9} \\ \hline
			W $\rightarrow$ C & 19.9 & 28.8 & 28.2 & 31.5 & 24.8 & 33.0 & 30.2 & 33.7 & 31.1 & 29.7 & 33.2 & \textbf{34.0} \\ \hline
			W $\rightarrow$ A & 23.0 & 27.6 & 29.1 & 30.5 & 27.6 & 29.8 & 30.0 & 36.0 & 30.0 & 38.3 & 39.9 & \textbf{42.7} \\ \hline
			W $\rightarrow$ D & 59.2 & 78.3 & 89.2 & 91.1 & 85.4 & \textbf{92.4} & 89.2 & 86.6 & 87.3 & 87.9 & 90.5 & 88.5 \\ \hline
			D $\rightarrow$ C & 26.3 & 26.4 & 29.7 & 33.0 & 29.3 & 31.2 & 31.4 & 33.8 & 30.7 & 30.5 & 29.9 & \textbf{34.9} \\ \hline
			D $\rightarrow$ A & 28.5 & 26.2 & 33.2 & 32.8 & 28.7 & 33.4 & 32.8 & 37.7 & 31.6 & 34.9 & 38.0 & \textbf{41.2} \\ \hline
			D $\rightarrow$ W & 63.4 & 52.5 & 86.1 & 87.5 & 80.3 & 89.2 & 85.4 & 84.7 & 84.4 & 88.5 & \textbf{91.9} & 87.5 \\ \hline \hline
			Average & 31.4 &  41.1 & 43.6 & 46.2 & 43.1 & 46.8 & 46.3 & 48.8 & 45.2 & 44.3 & 50.0 & \textbf{52.7} \\ \hline
		\end{tabular}
	}
\end{table*}

\begin{table*}[ht]
	\centering
	\vspace{-.1in}
	\caption{Accuracy (\%) on USPS+MNIST and ImageNet+VOC2007 datasets.}
	\label{tb-mnist}
	\vspace{-.15in}
	\resizebox{0.7\textwidth}{!}{
		\begin{tabular}{|c|c|c|c|c|c|c|c|c|c|c|c|c|c|}
			\hline
			Task & 1NN & SVM & PCA & TCA & GFK & JDA & TJM & CORAL & SCA & ARTL & JGSA & MEDA \\ \hline \hline
			U $\rightarrow$ M & 44.7 & 62.2 & 45.0 & 51.2 & 46.5 & 59.7 & 52.3 & 30.5 & 48.0 & 67.7 & 68.2 & \textbf{72.1} \\ \hline
			M $\rightarrow$ U & 65.9 & 68.2 & 66.2 & 56.3 & 61.2 & 67.3 & 63.3 & 49.2 & 65.1 & 88.8 & 80.4 & \textbf{89.5} \\ \hline
			I $\rightarrow$ V & 50.8 & 52.4 & 58.4 & 63.7 & 59.5 & 63.4 & 63.7 & 59.6 &-& 62.4 & 52.3 & \textbf{67.3} \\ \hline
			V $\rightarrow$ I & 38.2 & 42.7 & 65.1 & 64.9 & 73.8 & 70.2 & 73.0 & 70.3 & - &72.2 & 70.6 & \textbf{74.7} \\ \hline \hline
			Average & 49.9 & 56.3 & 58.7 & 59.0 & 60.2 & 65.1 & 63.1 & 52.4 & - & 72.8 & 67.9 & \textbf{75.9} \\ \hline
		\end{tabular}
	}
\end{table*}

\begin{table*}[ht]
	\centering
	\vspace{-.1in}
	\caption{Accuracy~(\%) on Office+Caltech10 datasets using DeCaf6 features.}
	\label{tb-decaf}
	\vspace{-.15in}
	\resizebox{1\textwidth}{!}{
		\begin{tabular}{|c|c|c|c|c|c|c|c|c|c|c|c|c|c|c|c|c|c|c|}
			\hline
			\multirow{2}{*}{Task} & \multicolumn{12}{c|}{Traditional Methods} & \multicolumn{5}{c|}{Deep Methods} & \multirow{2}{*}{MEDA} \\ \cline{2-18}
			& 1NN & SVM & PCA & TCA & GFK & JDA & TJM & SCA & ARTL & JGSA & CORAL & DMM & AlexNet & DDC & DAN & DCORAL & DUCDA &  \\ \hline \hline
			C $\rightarrow$ A & 87.3 & 91.6 & 88.1 & 89.8 & 88.2 & 89.6 & 88.8 & 89.5 & 92.4 & 91.4 & 92.0 & 92.4 & 91.9 & 91.9 & 92.0 & 92.4 & 92.8 & \textbf{93.4} \\ \hline
			C $\rightarrow$ W & 72.5 & 80.7 & 83.4 & 78.3 & 77.6 & 85.1 & 81.4 & 85.4 & 87.8 & 86.8 & 80.0 & 87.5 & 83.7 & 85.4 & 90.6 & 91.1 & 91.6 & \textbf{95.6} \\ \hline
			C $\rightarrow$ D & 79.6 & 86.0 & 84.1 & 85.4 & 86.6 & 89.8 & 84.7 & 87.9 & 86.6 & \textbf{93.6} & 84.7 & 90.4 & 87.1 & 88.8 & 89.3 & 91.4 & 91.7 & 91.1 \\ \hline
			A $\rightarrow$ C & 71.7 & 82.2 & 79.3 & 82.6 & 79.2 & 83.6 & 84.3 & 78.8 & 87.4 & 84.9 & 83.2 & 84.8 & 83.0 & 85.0 & 84.1 & 84.7 & 84.8 & \textbf{87.4} \\ \hline
			A $\rightarrow$ W & 68.1 & 71.9 & 70.9 & 74.2 & 70.9 & 78.3 & 71.9 & 75.9 & 88.5 & 81.0 & 74.6 & 84.7 & 79.5 & 86.1 & \textbf{91.8} & - & - & 88.1 \\ \hline
			A $\rightarrow$ D & 74.5 & 80.9 & 82.2 & 81.5 & 82.2 & 80.3 & 76.4 & 85.4 & 85.4 & 88.5 & 84.1 & \textbf{92.4} & 87.4 & 89.0 & 91.7 & - & - & 88.1 \\ \hline
			W $\rightarrow$ C & 55.3 & 67.9 & 70.3 & 80.4 & 69.8 & 84.8 & 83.0 & 74.8 & 88.2 & 85.0 & 75.5 & 81.7 & 73.0 & 78.0 & 81.2 & 79.3 & 80.2 & \textbf{93.2} \\ \hline
			W $\rightarrow$ A & 62.6 & 73.4 & 73.5 & 84.1 & 76.8 & 90.3 & 87.6 & 86.1 & 92.3 & 90.7 & 81.2 & 86.5 & 83.8 & 84.9 & 92.1 & - & - & \textbf{99.4} \\ \hline
			W $\rightarrow$ D & 98.1 & 100.0 & 99.4 & 100.0 & 100.0 & 100.0 & 100.0 & 100.0 & 100.0 & 100.0 & 100.0 & 98.7 & 100.0 & 100.0 & 100.0 & - & - & 99.4 \\ \hline
			D $\rightarrow$ C & 42.1 & 72.8 & 71.7 & 82.3 & 71.4 & 85.5 & 83.8 & 78.1 & 87.3 & 86.2 & 76.8 & 83.3 & 79.0 & 81.1 & 80.3 & 82.8 & 82.5 & \textbf{87.5} \\ \hline
			D $\rightarrow$ A & 50.0 & 78.7 & 79.2 & 89.1 & 76.3 & 91.7 & 90.3 & 90.0 & 92.7 & 92.0 & 85.5 & 90.7 & 87.1 & 89.5 & 90.0 & - & - & \textbf{93.2} \\ \hline
			D $\rightarrow$ W & 91.5 & 98.3 & 98.0 & 99.7 & 99.3 & 99.7 & 99.3 & 98.6 & \textbf{100.0} & 99.7 & 99.3 & 99.3 & 97.7 & 98.2 & 98.5 & - & - & 97.6 \\ \hline \hline
			Average & 71.1 & 82.0 & 81.7 & 85.6 & 81.5 & 88.2 & 86.0 & 85.9 & 90.7 & 90.0 & 84.7 & 89.4 & 86.1 & 88.2 & 90.1 & - & - & \textbf{92.8} \\ \hline	
		\end{tabular}
	}
\vspace{-.1in}
\end{table*}

\subsection{State-of-the-art Comparison Methods}
We compared the performance of MEDA with several state-of-the-art traditional and deep domain adaptation approaches.

Traditional learning methods:

\begin{itemize}[noitemsep,nolistsep]
	\item 1NN, SVM, and PCA
	\item Transfer Component Analysis~(\textbf{TCA})~\cite{pan2011domain}, which performs marginal distribution alignment
	\item Geodesic Flow Kernel~(\textbf{GFK})~\cite{gong2012geodesic}, which performs manifold feature learning
	\item Joint distribution alignment~(\textbf{JDA})~\cite{long2013transfer}, which adapts both marginal and conditional distribution
	\item Transfer Joint Matching~(\textbf{TJM})~\cite{long2014transfer}, which adapts marginal distribution with source sample selection
	\item Adaptation Regularization~(\textbf{ARTL})~\cite{long2014adaptation}, which learns domain classifier in original space
	\item CORrelation Alignment~(\textbf{CORAL})~\cite{sun2016return}, which performs second-order subspace alignment
	\item Scatter Component Analysis~(\textbf{SCA})~\cite{ghifary2017scatter}, which adapts scatters in subspace
	\item Joint Geometrical and Statistical Alignment~(\textbf{JGSA})~\cite{zhang2017joint}, which aligns marginal \& conditional distributions with label propagation
	\item Distribution Matching Machine~(\textbf{DMM})~\cite{cao2018unsupervised}, which learns a transfer SVM to align distributions
\end{itemize}

And deep domain adaptation methods:

\begin{itemize}[noitemsep,nolistsep]
	\item \textbf{AlexNet}~\cite{krizhevsky2012imagenet}, which is a standard convnet
	\item Deep Domain Confusion~(\textbf{DDC})~\cite{tzeng2014deep}, which is a single-layer deep adaptation method with MMD loss
	\item Deep Adaptation Network~(\textbf{DAN})~\cite{long2015domain}, which is a multi-layer adaptation method with multiple kernel MMD
	\item Deep CORAL (\textbf{DCORAL})~\cite{sun2016deep}, which is a deep neural network with CORAL loss
	\item Deep Unsupervised Convolutional Domain Adaptation (\textbf{DUCDA}) \cite{zhuo2017deep}, which is based on attention and CORAL loss
\end{itemize}

\subsection{Experimental Setup}

For fair comparison, we follow the same protocols as~\cite{zhang2017joint,zhuo2017deep,wang2017balanced} to adopt the extracted features for MEDA and other traditional methods. To be specific, 256 SURF features are used for USPS+MNIST datasets; for Office+Caltech10 datasets, both 800 SURF and 4,096 DeCaf6~\cite{donahue2014decaf} features are used; for Office-31 dataset, 4,096 DeCaf6 features are used; for ImageNet+VOC datasets, 4,096 DeCaf6 features are used. Deep methods can be used to the original images. 

Parameter setting: The optimal parameters of all comparison methods are set according to their original papers. As for MEDA, we set the manifold feature dimension $d=20,30,40$ for Office+Caltech10, USPS+MNIST, and ImageNet+VOC datasets, respectively. The iteration number are set to $T=10$. We use the RBF kernel with the bandwidth set to be the variance of inputs. The regularization parameters are set as $p=10,\lambda=10,\eta=0.1$, and $\rho=1$. The approach of setting these parameters are in the supplementary file. Additionally, the experiments on parameter sensitivity and convergence analysis in later experiments~(Section~\ref{sec-para} and~\ref{sec-conver}) indicate that MEDA~stays robust with a wide range of parameter choices.

We adopt classification Accuracy on $\mathcal{D}_t$ as the evaluation metric, which is widely used in existing literatures \cite{pan2011domain,wang2017balanced,gong2012geodesic}: $Accuracy = \frac{|\mathbf{x}: \mathbf{x} \in \mathcal{D}_t \wedge \hat{y}(\mathbf{x})=y(\mathbf{x})|}{|\mathbf{x}:\mathbf{x} \in \mathcal{D}_t|}$, where $y(\mathbf{x})$ and $\hat{y}(\mathbf{x})$ are the truth and predicted labels for target domain, respectively.

\begin{figure}[t!]
	\centering
	\vspace{-.1in}
	\includegraphics[scale=0.3]{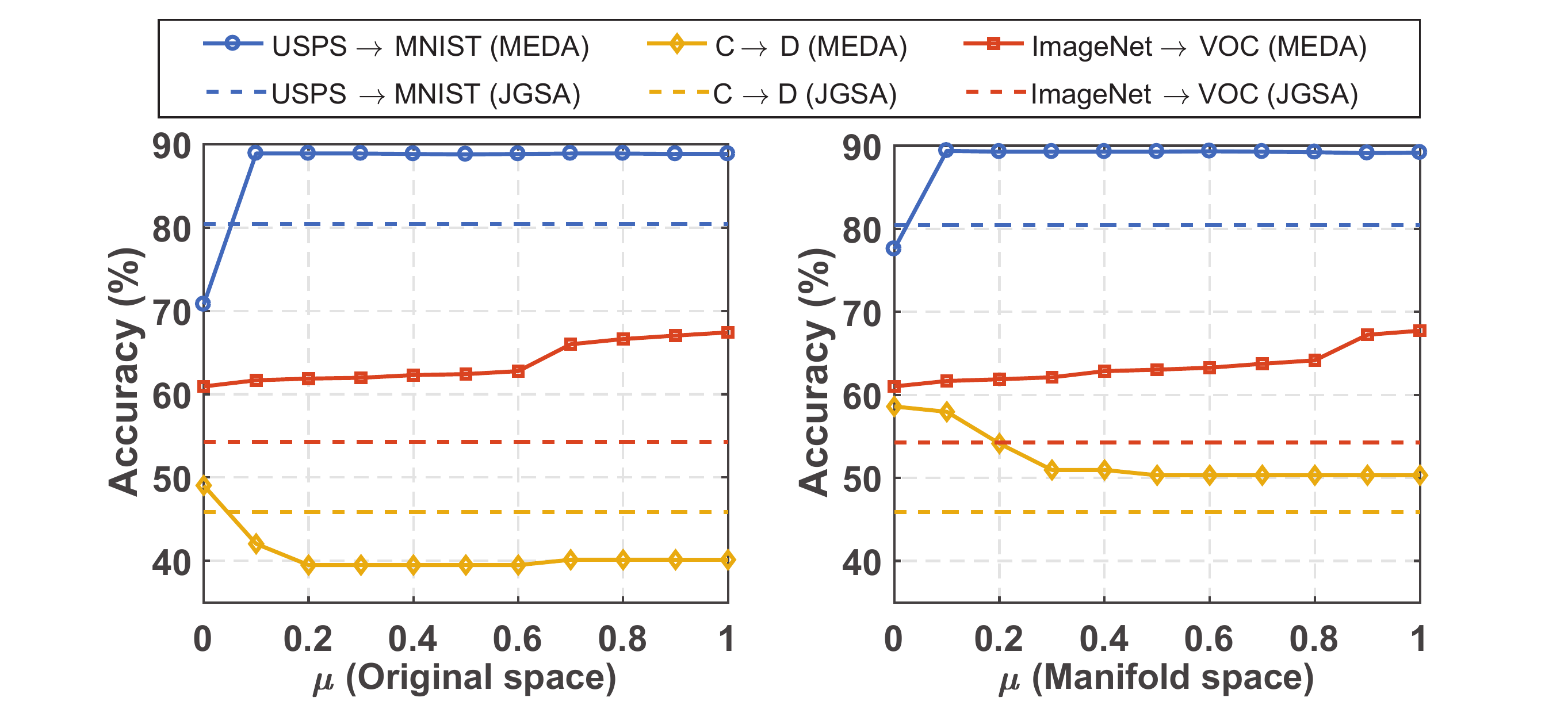}
	\vspace{-.1in}
	\caption{Accuracy in original (left) and manifold space (right) with different $\mu$. Dashed lines are best baseline.}
	\label{fig-mu}
	\vspace{-.1in}
\end{figure}

\begin{figure}[t!]
	\centering
	\vspace{-.1in}
	\includegraphics[scale=0.4]{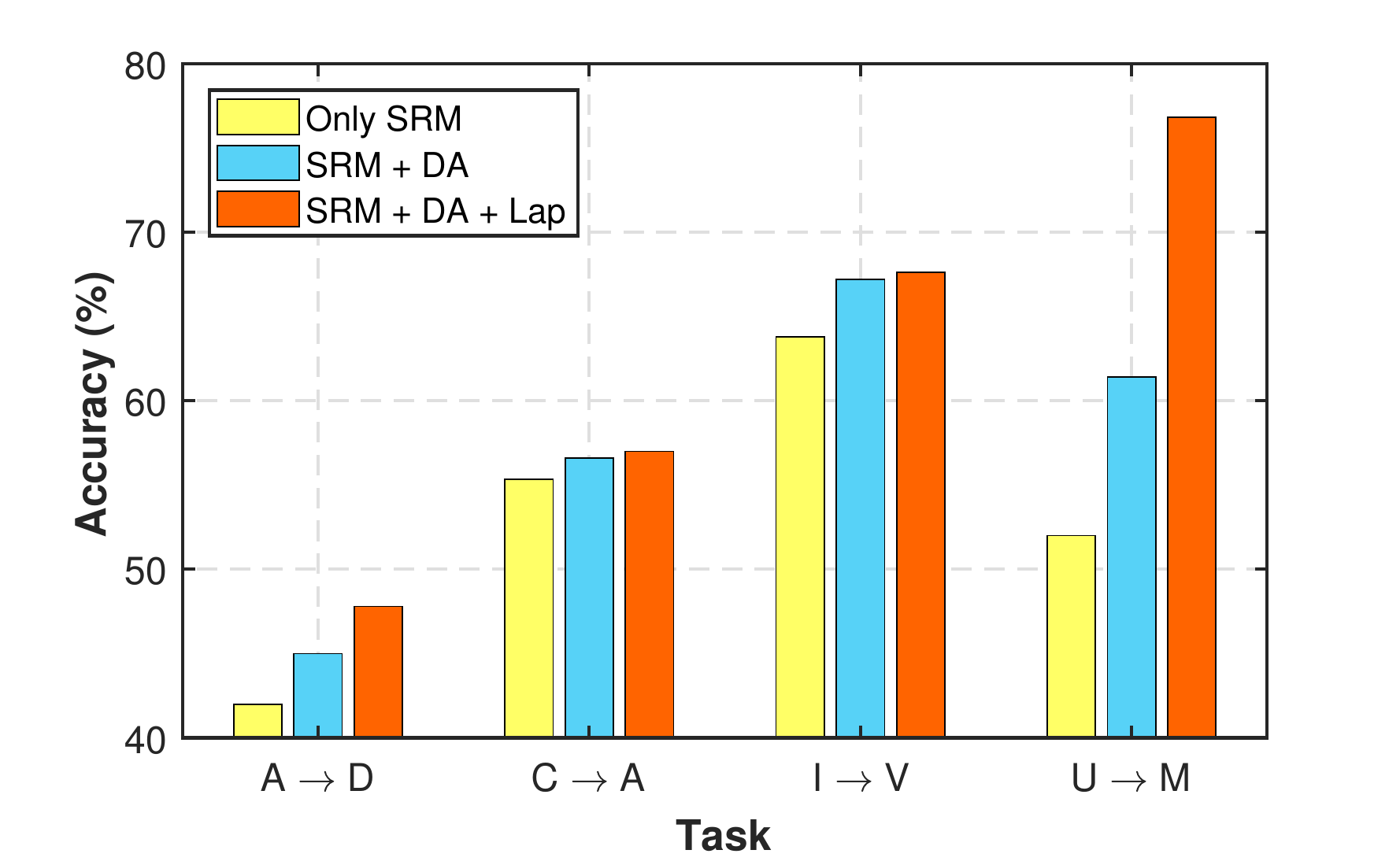}
	\vspace{-.15in}
	\caption{Evaluation of each component.}
	\label{fig-component}
	\vspace{-.15in}
\end{figure}

\subsection{Experimental Results and Analysis}
The classification accuracy results on the aforementioned datasets are shown in Tables~\ref{tb-surf}, \ref{tb-mnist}, and \ref{tb-decaf}, respectively\footnote{Symbol `-' denotes the result is not available since there is no code or results.}. From those results, we can make several observations as follows. 

Firstly, MEDA outperformed all other traditional and deep comparison methods in most tasks (21/28 tasks). The average classification accuracy of MEDA on 28 tasks was $\bm{73.2}$\textbf{\%}. Compared to the best baseline method JGSA (69.7\%), the average performance improvement was $\bm{3.5}$\textbf{\%}, which showed a significant average error reduction of $\bm{11.6}$\textbf{\%}. Note that the results on Office-31 dataset were in the supplementary file~\footnote{Supplementary file is at \url{https://www.jianguoyun.com/p/DRuWOFkQjKnsBRjkr2E}.} due to space constraints, and the observations are the same. Since these results were obtained from a wide range of image datasets, it demonstrates that MEDA is capable of significantly reducing the distribution divergence in domain adaptation problems. 

Secondly, the performances of distribution alignment methods (TCA, JDA, ARTL, TJM, JGSA, and DMM) and subspace learning methods~(GFK, CORAL, and SCA) were generally worse than MEDA. Each kind of methods has its limitations and cannot handle domain adaptation in specific tasks. This indicates the disadvantages of those methods to cope with degenerated feature transformation and unevaluated distribution alignment. After manifold or subsapce learning, there still exists large domain shift~\cite{baktashmotlagh2013unsupervised}; while feature distortion will undermine the distribution alignment methods. 

Thirdly, MEDA also outperformed the deep methods (AlexNet, DDC, DAN, DCORAL, and DUCDA) on Office+Caltech10 datasets. Deep methods often have to tune a lot of hyperparameters before obtaining the optimal results. Compared to them, MEDA only involves several parameters that can easily be set by human experience or cross-validation. This implies the accuracy and efficiency of MEDA in domain adaptation problems over other deep methods.


\begin{table}[t!]
	\centering
	\vspace{-.1in}
	\caption{Mean and standard deviation of accuracy in feature learning in both original and manifold space.}
	\vspace{-.15in}
	\label{tb-gfk}
	\resizebox{0.48\textwidth}{!}{
		\begin{tabular}{|c|c|c|r|}
			\hline
			\textbf{Task} & \textbf{Original Space} & \textbf{Manifold Space} & \multicolumn{1}{c|}{\textbf{Improvement}} \\ \hline
			C $\rightarrow$ A & 44.9 (2.1) & 56.5 (0.5) & \textbf{25.8\% (-76.9\%)} \\ \hline
			C $\rightarrow$ W & 33.5 (4.5) & 54.0 (0.4) & \textbf{61.4\% (-90.9\%)} \\ \hline
			C $\rightarrow$ A (DeCaf) & 92.5 (0.2) & 93.4 (0.1) & \textbf{1.0\% (-58.3\%)} \\ \hline
			C $\rightarrow$ W (DeCaf) & 88.4 (1.7) & 95.5 (0.3) & \textbf{8.1\% (-82.6\%)} \\ \hline
			U $\rightarrow$ M & 64.1 (9.2) & 71.2 (4.2) & \textbf{11.1\% (-54.5\%)} \\ \hline
			I $\rightarrow$ V & 63.0 (2.5) & 63.7 (2.2) & \textbf{1.1\% (-13.2\%)} \\ \hline
		\end{tabular}
	}
\end{table}

\begin{table}[t!]
	\centering
	\vspace{-.1in}
	\caption{Performance comparison between $\mu_{opt}$ and $\hat{\mu}$.}
	\label{tb-mu}
	\vspace{-.15in}
	\resizebox{.5\textwidth}{!}{
		\begin{tabular}{|m{0.18\columnwidth}<{\centering}|c|c|c|c|c|c|}
			\hline
			\textbf{Task} & \textbf{C $\rightarrow$ A} & \textbf{W $\rightarrow$ D} & \textbf{C $\rightarrow$ A (DeCaf)} & \textbf{W $\rightarrow$ C (DeCaf)} & \textbf{M $\rightarrow$ U} & \textbf{I $\rightarrow$ V} \\ \hline
			$\mu_{opt}$ & 57.0 & 89.2 & 93.4 & 88.0 & 89.4 & 67.6 \\ \hline
			$\hat{\mu}$ & 56.5 & 88.5 & 93.4 & 93.2 & 89.5 & 67.3 \\ \hline
			Performance Variation & -0.9\% & -0.8\% & 0 & \textbf{+5.9}\% & \textbf{+0.1}\% & -0.4\% \\ \hline
		\end{tabular}
	}
\vspace{-.2in}
\end{table}

\subsection{Effectiveness Analysis}

\subsubsection{Manifold Feature Learning} 
\label{sec-mani}
We investigate the effectiveness of manifold feature learning in handling the degenerated feature transformation challenge. To this end, we ran MEDA with and without manifold feature learning on randomly selected tasks. Table~\ref{tb-gfk} showed the mean, standard deviation, and performance improvement of classification accuracy with $\mu \in \{0,0.1,\cdots,1\}$. For instance, the improvement of mean accuracy on task C $\rightarrow$ A was: $(56.5-44.9)/44.9 \times 100\%=25.8\%$. From these results, we can observe that: 1) The performance of all the tasks were improved with manifold feature learning, indicating that transforming features into the manifold alleviates domain shift to some extent and facilitates distribution alignment; 2) The standard deviation of methods that adopted manifold learning with different $\mu$ could be dramatically reduced. 3) MEDA can also reach a comparable performance without manifold learning, while adding manifold learning would produce better results. This reveals the effectiveness of manifold feature learning to alleviate degenerated feature transformation. 

\begin{figure*}[t!]
	\centering
	\vspace{-.15in}
	\hspace{-.2in}
	\subfigure[Subspace dimension $d$]{
		\includegraphics[scale=0.28]{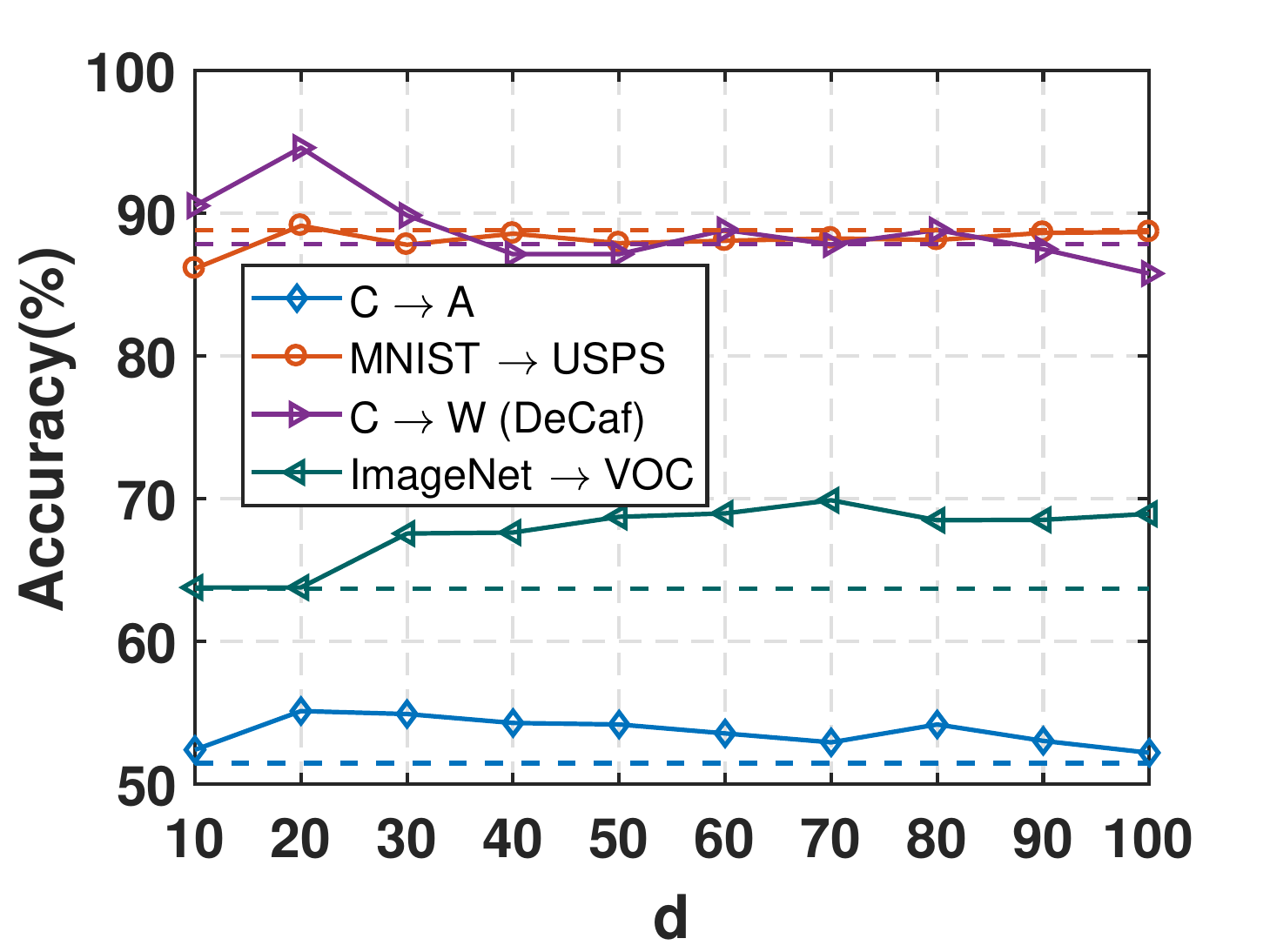}
		\label{fig-sub-d}}
	\subfigure[\#neighbor $p$]{
		\includegraphics[scale=0.28]{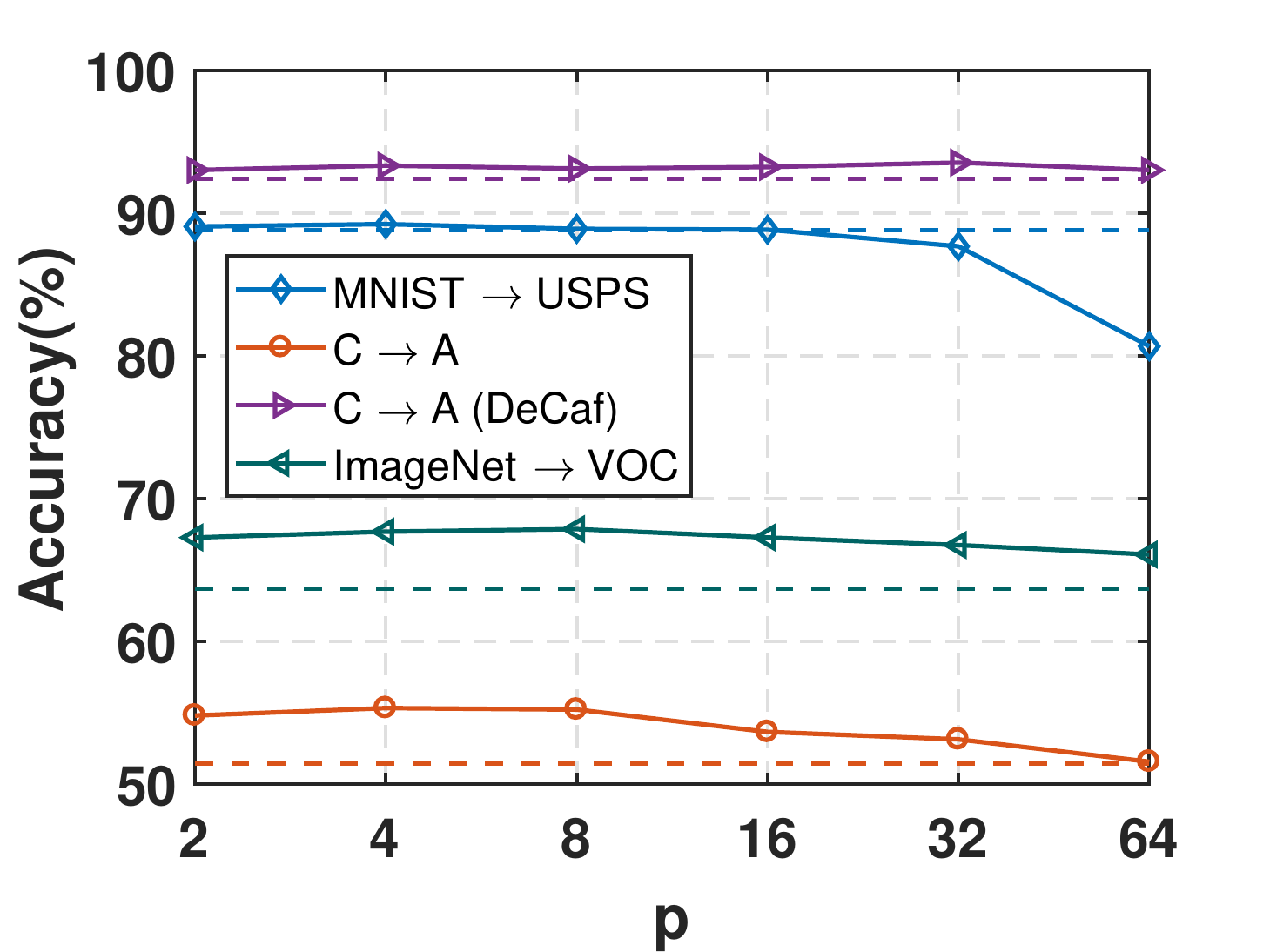}
		\label{fig-sub-p}}
	\subfigure[$\lambda$]{
		\includegraphics[scale=0.28]{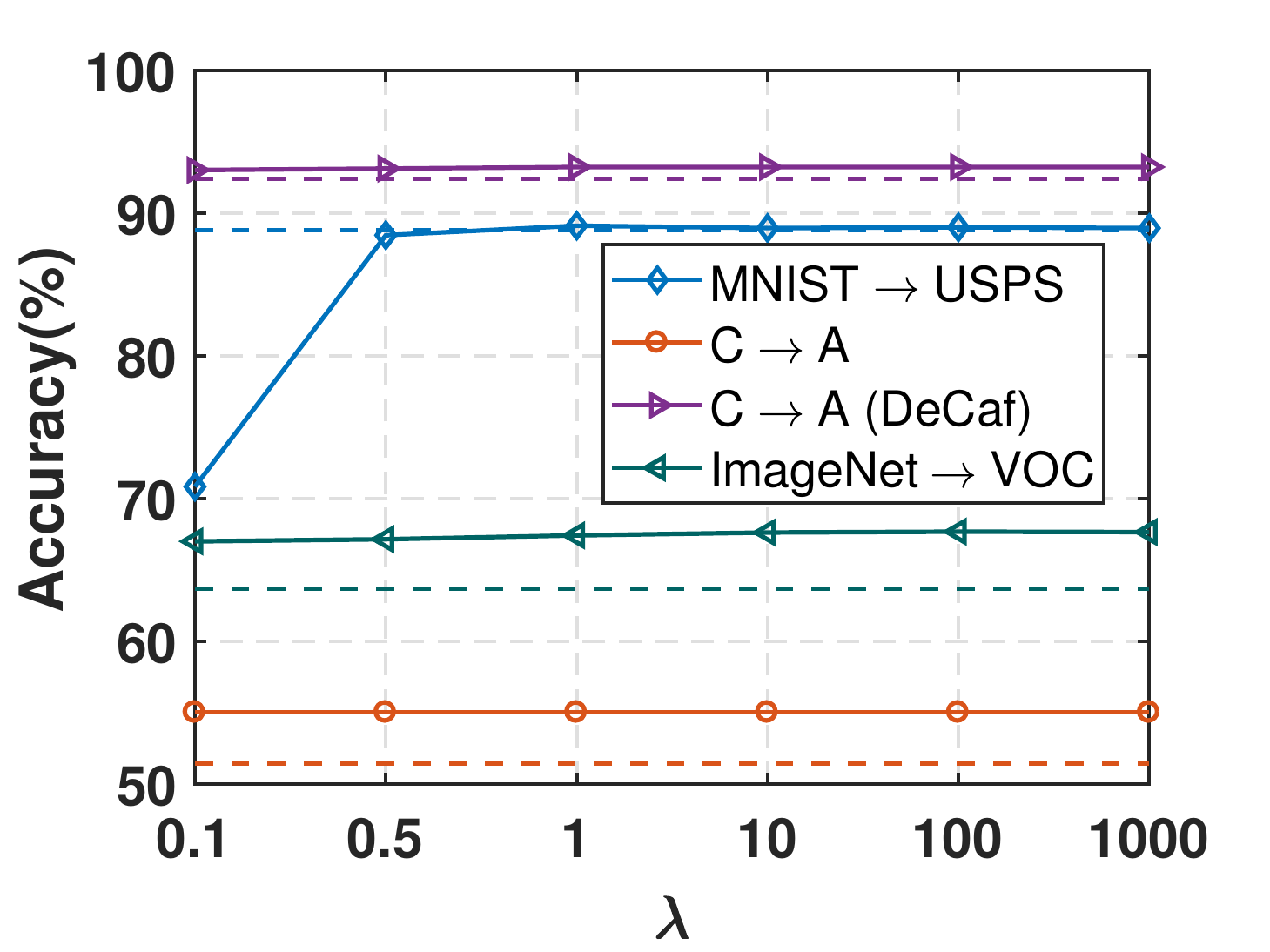}
		\label{fig-sub-lambda}}
	\subfigure[Convergence]{
		\includegraphics[scale=0.28]{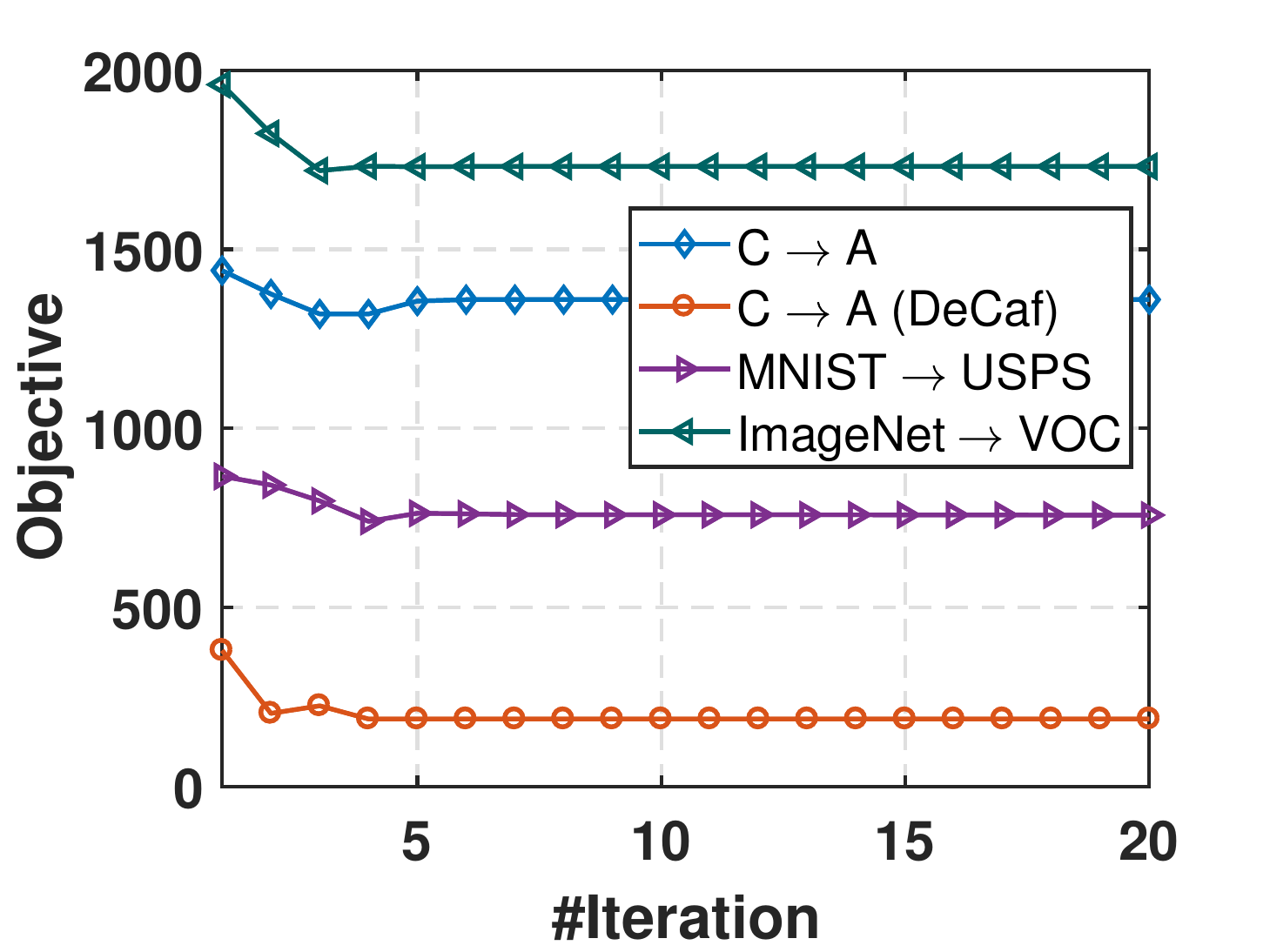}
		\label{fig-sub-iteration}}
	\vspace{-.15in}
	\caption{(a)$\sim$(c): classification accuracy w.r.t. $d$, $p$, and $\lambda$, respectively. (d)~convergence analysis.}
	\vspace{-.15in}
	\label{fig-p-d}
\end{figure*}

\subsubsection{Dynamic Distribution Alignment}

We verify the effectiveness of dynamic distribution alignment in handling the \textit{unevaluated distribution alignment} challenge. We ran MEDA by searching $\mu \in \{0,0.1,\cdots,0.9,1.0\}$ and compared the performances with the best baseline method (JGSA). From the results in Figure~\ref{fig-mu}, we can clearly observe that the classification accuracy varied with different choice of $\mu$. This indicates the \textit{necessity} to consider the different effects between marginal and conditional distributions. We can also observe that the optimal $\mu$ value varied on different tasks ($\mu=0.2,0,1$ for three tasks, respectively). Thus, it is necessary to dynamically adjust the distribution alignment between domains according to different tasks. Moreover, the optimal value of $\mu$ is \textit{not} unique on certain task. The classification results may be the same even for different $\mu$.

\textit{The estimation of } $\mu$: We evaluate our solution of estimating $\mu$~(equation~(\ref{eq-mu})). Since the optimal $\mu$ is not unique, we can not directly compare the value of $\mu_{opt}$ and $\hat{\mu}$ to evaluate our solution. Instead, we compare the performances (accuracy values) achieved by $\mu_{opt}$ and $\hat{\mu}$. The results in Table~\ref{tb-mu} indicated that the performance of estimated $\hat{\mu}$ was very close to $\mu_{opt}$, and sometimes it is better than grid search (M $\rightarrow$ U). For instance, the performance variation of C $\rightarrow$ A was $(57.0-56.5) / 57.0 \times 100\% = 0.9\%$. This demonstrates that the effectiveness in estimating $\mu$. This estimation solution can be directly applied to future research.

\subsubsection{Evaluation of Each Component} When learning the final classifier $f$, MEDA involves three components: the structural risk minimization (SRM), the dynamic distribution alignment (DA), and Laplacian regularization (Lap). We empirically evaluated the importance of each component. We randomly selected several tasks and reported the results in Figure~\ref{fig-component}. Note that we did not run this experiment on the Decaf features of Office+Caltech10 dataset since its results are already satisfied.

Those results clearly indicated that each component is important in MEDA, and they are indispensable. Moreover, we observe that in all tasks, it is more important to align the \textit{distributions}. The reason is that there exists large distribution divergence between two domains. The results also suggests that adding Laplacian regularization is more beneficial in capturing the manifold structure. Additionally, combining the effectiveness of manifold feature learning~(Section~\ref{sec-mani}), it is clear that all components are important for improving the accuracy in domain adaptation tasks.

\subsection{Parameter Sensitivity}
\label{sec-para}

As with other state-of-the-art domain adaptation algorithms~\cite{zhang2017joint,long2014adaptation,ghifary2017scatter}, MEDA also involves several parameters. In this section, we evaluate the parameter sensitivity. Due to lack of space, we only report the main results in this paper. Other results can be found in the supplementary file. Experimental results demonstrated the robustness of MEDA under a wide range of parameter choices. Therefore, the parameters do not need to be fine-tuned in real applications.

\subsubsection{Subspace Dimension and \#neighbor}

We investigated the sensitivity of manifold subspace dimension $d$ and \#neighbor $p$ through experiments with a wide range of $d \in \{10,20,\cdots,100\}$ and $p \in \{2,4,\cdots,64\}$ on randomly selected tasks. From the results in Figure~\ref{fig-sub-d} and \ref{fig-sub-p}, it can be observed that MEDA was robust with regard to different values of $d$ and $p$. Therefore, they can be selected without knowledge in real applications. 

\subsubsection{Regularization Parameters}

We ran MEDA with a wide range of values for regularization parameters $\lambda,\eta$, and $\rho$ on several random tasks and compare its performance with the best baseline method. For the lack of space, we only report the results of $\lambda$ in Figure~\ref{fig-sub-lambda}, and the results of $\rho$ and $\eta$ can be found in the supplementary file. We observed that MEDA can achieve a robust performance with regard to a wide range of parameter values. Specifically, the best choices of these parameters are: $\lambda \in [0.5, 1,000], \eta \in [0.01,1]$, and $\rho \in [0.01,5]$. To sum up, the performance of MEDA stays robust with a wide range of regularization parameter choice. 

\subsection{Convergence and Time Complexity}
\label{sec-conver}

We validated the convergence of MEDA through empirical analysis. From the results in Figure~\ref{fig-sub-iteration}, it can be observed that MEDA can reach a steady performance in only a few $(T < 10)$ iterations. It indicates the training advantage of MEDA in cross-domain tasks.

We also empirically checked the time complexity of MEDA and compared it with other top two baselines ARTL and JGSA on different tasks. The environment was an Intel Core i7-4790 CPU with 24 GB memory. Note that the time complexity of deep methods are not comparable with MEDA since they require a lot of backpropagations. The results in Table~\ref{tb-time} reveal that except its superiority in classification accuracy, MEDA also achieved a running time complexity comparable to top two best baseline methods.

\begin{table}[]
	\centering
	\caption{Running time~(s) of ARTL, JGSA, and MEDA.}
	\label{tb-time}
	\vspace{-.15in}
	\resizebox{0.48\textwidth}{!}{
		\begin{tabular}{|c|c|c|c|c|}
			\hline
			\multicolumn{1}{|c|}{\textbf{Task}} & \multicolumn{1}{|c|}{\textbf{\#Sample} $\times$ \textbf{\#Feature}} & \multicolumn{1}{c|}{\textbf{ARTL}} & \multicolumn{1}{c|}{\textbf{JGSA}} & \multicolumn{1}{c|}{\textbf{MEDA}} \\ \hline \hline
			C $\rightarrow$ A & 2,081 $\times$ 800 & 29.2 & 95.2 & 32.3 \\ \hline
			M $\rightarrow$ U & 3,800 $\times$ 256 & 29.1 & 14.6 & 31.4 \\ \hline
			I $\rightarrow$ V & 10,717 $\times$ 4,096 & 2,648.8 & $>$ 10,000 & 2,931.7 \\ \hline
		\end{tabular}
	}
\vspace{-.2in}
\end{table}

\label{sec-con}
\section{Conclusions}

In this paper, we propose a novel Manifold Embedded Distribution Alignment~(MEDA) approach for visual domain adaptation. Compared to existing work, MEDA is the first attempt to handle the challenges of both degenerated feature transformation and unevaluated distribution alignment. MEDA can learn the domain-invariant classifier with the principle of structural risk minimization while performing dynamic distribution alignment. We also provide a feasible solution to quantitatively calculate the adaptive factor. We conducted extensive experiments on several large-scale publicly available image classification datasets. The results demonstrate the superiority of MEDA against other state-of-the-art traditional and deep domain adaptation methods. 


\begin{acks}
	This work is supported in part by National Key R \& D Plan of China (2016YFB1001200), NSFC
	(61572471,61702520,61672313), and NSF through grants IIS-1526499, IIS-1763325, CNS-1626432, and Nanyang Assistant Professorship (NAP) of Nanyang Technological University.
	
\end{acks}

\bibliographystyle{ACM-Reference-Format}
\bibliography{mm18}


\begin{thebibliography}{41}


\ifx \showCODEN    \undefined \def \showCODEN     #1{\unskip}     \fi
\ifx \showDOI      \undefined \def \showDOI       #1{#1}\fi
\ifx \showISBNx    \undefined \def \showISBNx     #1{\unskip}     \fi
\ifx \showISBNxiii \undefined \def \showISBNxiii  #1{\unskip}     \fi
\ifx \showISSN     \undefined \def \showISSN      #1{\unskip}     \fi
\ifx \showLCCN     \undefined \def \showLCCN      #1{\unskip}     \fi
\ifx \shownote     \undefined \def \shownote      #1{#1}          \fi
\ifx \showarticletitle \undefined \def \showarticletitle #1{#1}   \fi
\ifx \showURL      \undefined \def \showURL       {\relax}        \fi
\providecommand\bibfield[2]{#2}
\providecommand\bibinfo[2]{#2}
\providecommand\natexlab[1]{#1}
\providecommand\showeprint[2][]{arXiv:#2}

\bibitem[\protect\citeauthoryear{Aljundi, Emonet, Muselet, and Sebban}{Aljundi
  et~al\mbox{.}}{2015}]%
        {aljundi2015landmarks}
\bibfield{author}{\bibinfo{person}{Rahaf Aljundi}, \bibinfo{person}{R{\'e}mi
  Emonet}, \bibinfo{person}{Damien Muselet}, {and} \bibinfo{person}{Marc
  Sebban}.} \bibinfo{year}{2015}\natexlab{}.
\newblock \showarticletitle{Landmarks-based kernelized subspace alignment for
  unsupervised domain adaptation}. In \bibinfo{booktitle}{\emph{Proceedings of
  the IEEE Conference on Computer Vision and Pattern Recognition}}.
  \bibinfo{pages}{56--63}.
\newblock


\bibitem[\protect\citeauthoryear{Baktashmotlagh, Harandi, and
  Salzmann}{Baktashmotlagh et~al\mbox{.}}{2016}]%
        {baktashmotlagh2016distribution}
\bibfield{author}{\bibinfo{person}{Mahsa Baktashmotlagh},
  \bibinfo{person}{Mehrtash Harandi}, {and} \bibinfo{person}{Mathieu
  Salzmann}.} \bibinfo{year}{2016}\natexlab{}.
\newblock \showarticletitle{Distribution-matching embedding for visual domain
  adaptation}.
\newblock \bibinfo{journal}{\emph{The Journal of Machine Learning Research}}
  \bibinfo{volume}{17}, \bibinfo{number}{1} (\bibinfo{year}{2016}),
  \bibinfo{pages}{3760--3789}.
\newblock


\bibitem[\protect\citeauthoryear{Baktashmotlagh, Harandi, Lovell, and
  Salzmann}{Baktashmotlagh et~al\mbox{.}}{2013}]%
        {baktashmotlagh2013unsupervised}
\bibfield{author}{\bibinfo{person}{Mahsa Baktashmotlagh},
  \bibinfo{person}{Mehrtash~T Harandi}, \bibinfo{person}{Brian~C Lovell}, {and}
  \bibinfo{person}{Mathieu Salzmann}.} \bibinfo{year}{2013}\natexlab{}.
\newblock \showarticletitle{Unsupervised domain adaptation by domain invariant
  projection}. In \bibinfo{booktitle}{\emph{Proceedings of the IEEE
  International Conference on Computer Vision}}. \bibinfo{pages}{769--776}.
\newblock


\bibitem[\protect\citeauthoryear{Baktashmotlagh, Harandi, Lovell, and
  Salzmann}{Baktashmotlagh et~al\mbox{.}}{2014}]%
        {baktashmotlagh2014domain}
\bibfield{author}{\bibinfo{person}{Mahsa Baktashmotlagh},
  \bibinfo{person}{Mehrtash~T Harandi}, \bibinfo{person}{Brian~C Lovell}, {and}
  \bibinfo{person}{Mathieu Salzmann}.} \bibinfo{year}{2014}\natexlab{}.
\newblock \showarticletitle{Domain adaptation on the statistical manifold}. In
  \bibinfo{booktitle}{\emph{Proceedings of the IEEE Conference on Computer
  Vision and Pattern Recognition}}. \bibinfo{pages}{2481--2488}.
\newblock


\bibitem[\protect\citeauthoryear{Belkin, Niyogi, and Sindhwani}{Belkin
  et~al\mbox{.}}{2006}]%
        {belkin2006manifold}
\bibfield{author}{\bibinfo{person}{Mikhail Belkin}, \bibinfo{person}{Partha
  Niyogi}, {and} \bibinfo{person}{Vikas Sindhwani}.}
  \bibinfo{year}{2006}\natexlab{}.
\newblock \showarticletitle{Manifold regularization: A geometric framework for
  learning from labeled and unlabeled examples}.
\newblock \bibinfo{journal}{\emph{Journal of machine learning research}}
  \bibinfo{volume}{7}, \bibinfo{number}{Nov} (\bibinfo{year}{2006}),
  \bibinfo{pages}{2399--2434}.
\newblock


\bibitem[\protect\citeauthoryear{Ben-David, Blitzer, Crammer, and
  Pereira}{Ben-David et~al\mbox{.}}{2007}]%
        {ben2007analysis}
\bibfield{author}{\bibinfo{person}{Shai Ben-David}, \bibinfo{person}{John
  Blitzer}, \bibinfo{person}{Koby Crammer}, {and} \bibinfo{person}{Fernando
  Pereira}.} \bibinfo{year}{2007}\natexlab{}.
\newblock \showarticletitle{Analysis of representations for domain adaptation}.
  In \bibinfo{booktitle}{\emph{Advances in neural information processing
  systems}}. \bibinfo{pages}{137--144}.
\newblock


\bibitem[\protect\citeauthoryear{Cao, Long, and Wang}{Cao
  et~al\mbox{.}}{2018}]%
        {cao2018unsupervised}
\bibfield{author}{\bibinfo{person}{Yue Cao}, \bibinfo{person}{Mingsheng Long},
  {and} \bibinfo{person}{Jianmin Wang}.} \bibinfo{year}{2018}\natexlab{}.
\newblock \showarticletitle{Unsupervised Domain Adaptation with Distribution
  Matching Machines}. In \bibinfo{booktitle}{\emph{Proceedings of the 2018 AAAI
  International Conference on Artificial Intelligence}}.
\newblock


\bibitem[\protect\citeauthoryear{Chen, Zhang, Xiao, Liu, and Chang}{Chen
  et~al\mbox{.}}{2018}]%
        {chen2018zero}
\bibfield{author}{\bibinfo{person}{Long Chen}, \bibinfo{person}{Hanwang Zhang},
  \bibinfo{person}{Jun Xiao}, \bibinfo{person}{Wei Liu}, {and}
  \bibinfo{person}{Shih-Fu Chang}.} \bibinfo{year}{2018}\natexlab{}.
\newblock \showarticletitle{Zero-Shot Visual Recognition using
  Semantics-Preserving Adversarial Embedding Network}. In
  \bibinfo{booktitle}{\emph{IEEE Conference on Computer Vision and Pattern
  Recognition (CVPR)}}.
\newblock


\bibitem[\protect\citeauthoryear{Dai, Yang, Xue, and Yu}{Dai
  et~al\mbox{.}}{2007}]%
        {dai2007boosting}
\bibfield{author}{\bibinfo{person}{Wenyuan Dai}, \bibinfo{person}{Qiang Yang},
  \bibinfo{person}{Gui-Rong Xue}, {and} \bibinfo{person}{Yong Yu}.}
  \bibinfo{year}{2007}\natexlab{}.
\newblock \showarticletitle{Boosting for transfer learning}. In
  \bibinfo{booktitle}{\emph{Proceedings of the 24th international conference on
  Machine learning (ICML)}}. ACM, \bibinfo{pages}{193--200}.
\newblock


\bibitem[\protect\citeauthoryear{Denman and Beavers~Jr}{Denman and
  Beavers~Jr}{1976}]%
        {denman1976matrix}
\bibfield{author}{\bibinfo{person}{Eugene~D Denman} {and}
  \bibinfo{person}{Alex~N Beavers~Jr}.} \bibinfo{year}{1976}\natexlab{}.
\newblock \showarticletitle{The matrix sign function and computations in
  systems}.
\newblock \bibinfo{journal}{\emph{Applied mathematics and Computation}}
  \bibinfo{volume}{2}, \bibinfo{number}{1} (\bibinfo{year}{1976}),
  \bibinfo{pages}{63--94}.
\newblock


\bibitem[\protect\citeauthoryear{Donahue, Jia, Vinyals, Hoffman, Zhang, Tzeng,
  and Darrell}{Donahue et~al\mbox{.}}{2014}]%
        {donahue2014decaf}
\bibfield{author}{\bibinfo{person}{Jeff Donahue}, \bibinfo{person}{Yangqing
  Jia}, \bibinfo{person}{Oriol Vinyals}, \bibinfo{person}{Judy Hoffman},
  \bibinfo{person}{Ning Zhang}, \bibinfo{person}{Eric Tzeng}, {and}
  \bibinfo{person}{Trevor Darrell}.} \bibinfo{year}{2014}\natexlab{}.
\newblock \showarticletitle{Decaf: A deep convolutional activation feature for
  generic visual recognition}. In \bibinfo{booktitle}{\emph{International
  conference on machine learning}}. \bibinfo{pages}{647--655}.
\newblock


\bibitem[\protect\citeauthoryear{Fang, Xu, and Rockmore}{Fang
  et~al\mbox{.}}{2013}]%
        {fang2013unbiased}
\bibfield{author}{\bibinfo{person}{Chen Fang}, \bibinfo{person}{Ye Xu}, {and}
  \bibinfo{person}{Daniel~N Rockmore}.} \bibinfo{year}{2013}\natexlab{}.
\newblock \showarticletitle{Unbiased metric learning: On the utilization of
  multiple datasets and web images for softening bias}. In
  \bibinfo{booktitle}{\emph{Proceedings of the IEEE International Conference on
  Computer Vision}}. \bibinfo{pages}{1657--1664}.
\newblock


\bibitem[\protect\citeauthoryear{Fernando, Habrard, Sebban, and
  Tuytelaars}{Fernando et~al\mbox{.}}{2013}]%
        {fernando2013unsupervised}
\bibfield{author}{\bibinfo{person}{Basura Fernando}, \bibinfo{person}{Amaury
  Habrard}, \bibinfo{person}{Marc Sebban}, {and} \bibinfo{person}{Tinne
  Tuytelaars}.} \bibinfo{year}{2013}\natexlab{}.
\newblock \showarticletitle{Unsupervised visual domain adaptation using
  subspace alignment}. In \bibinfo{booktitle}{\emph{Proceedings of the IEEE
  international conference on computer vision}}. \bibinfo{pages}{2960--2967}.
\newblock


\bibitem[\protect\citeauthoryear{Ghifary, Balduzzi, Kleijn, and Zhang}{Ghifary
  et~al\mbox{.}}{2017}]%
        {ghifary2017scatter}
\bibfield{author}{\bibinfo{person}{Muhammad Ghifary}, \bibinfo{person}{David
  Balduzzi}, \bibinfo{person}{W~Bastiaan Kleijn}, {and}
  \bibinfo{person}{Mengjie Zhang}.} \bibinfo{year}{2017}\natexlab{}.
\newblock \showarticletitle{Scatter component analysis: A unified framework for
  domain adaptation and domain generalization}.
\newblock \bibinfo{journal}{\emph{IEEE transactions on pattern analysis and
  machine intelligence}} \bibinfo{volume}{39}, \bibinfo{number}{7}
  (\bibinfo{year}{2017}), \bibinfo{pages}{1414--1430}.
\newblock


\bibitem[\protect\citeauthoryear{Gong, Shi, Sha, and Grauman}{Gong
  et~al\mbox{.}}{2012}]%
        {gong2012geodesic}
\bibfield{author}{\bibinfo{person}{Boqing Gong}, \bibinfo{person}{Yuan Shi},
  \bibinfo{person}{Fei Sha}, {and} \bibinfo{person}{Kristen Grauman}.}
  \bibinfo{year}{2012}\natexlab{}.
\newblock \showarticletitle{Geodesic flow kernel for unsupervised domain
  adaptation}. In \bibinfo{booktitle}{\emph{Computer Vision and Pattern
  Recognition (CVPR), 2012 IEEE Conference on}}. IEEE,
  \bibinfo{pages}{2066--2073}.
\newblock


\bibitem[\protect\citeauthoryear{Gopalan, Li, and Chellappa}{Gopalan
  et~al\mbox{.}}{2011}]%
        {gopalan2011domain}
\bibfield{author}{\bibinfo{person}{Raghuraman Gopalan}, \bibinfo{person}{Ruonan
  Li}, {and} \bibinfo{person}{Rama Chellappa}.}
  \bibinfo{year}{2011}\natexlab{}.
\newblock \showarticletitle{Domain adaptation for object recognition: An
  unsupervised approach}. In \bibinfo{booktitle}{\emph{Computer Vision (ICCV),
  2011 IEEE International Conference on}}. IEEE, \bibinfo{pages}{999--1006}.
\newblock


\bibitem[\protect\citeauthoryear{Gretton, Borgwardt, Rasch, Sch{\"o}lkopf, and
  Smola}{Gretton et~al\mbox{.}}{2012}]%
        {gretton2012kernel}
\bibfield{author}{\bibinfo{person}{Arthur Gretton}, \bibinfo{person}{Karsten~M
  Borgwardt}, \bibinfo{person}{Malte~J Rasch}, \bibinfo{person}{Bernhard
  Sch{\"o}lkopf}, {and} \bibinfo{person}{Alexander Smola}.}
  \bibinfo{year}{2012}\natexlab{}.
\newblock \showarticletitle{A kernel two-sample test}.
\newblock \bibinfo{journal}{\emph{Journal of Machine Learning Research}}
  \bibinfo{volume}{13}, \bibinfo{number}{Mar} (\bibinfo{year}{2012}),
  \bibinfo{pages}{723--773}.
\newblock


\bibitem[\protect\citeauthoryear{Hamm and Lee}{Hamm and Lee}{2008}]%
        {hamm2008grassmann}
\bibfield{author}{\bibinfo{person}{Jihun Hamm} {and} \bibinfo{person}{Daniel~D
  Lee}.} \bibinfo{year}{2008}\natexlab{}.
\newblock \showarticletitle{Grassmann discriminant analysis: a unifying view on
  subspace-based learning}. In \bibinfo{booktitle}{\emph{Proceedings of the
  25th international conference on Machine learning}}. ACM,
  \bibinfo{pages}{376--383}.
\newblock


\bibitem[\protect\citeauthoryear{Hou, Tsai, Yeh, and Wang}{Hou
  et~al\mbox{.}}{2016}]%
        {hou2016unsupervised}
\bibfield{author}{\bibinfo{person}{Cheng-An Hou},
  \bibinfo{person}{Yao-Hung~Hubert Tsai}, \bibinfo{person}{Yi-Ren Yeh}, {and}
  \bibinfo{person}{Yu-Chiang~Frank Wang}.} \bibinfo{year}{2016}\natexlab{}.
\newblock \showarticletitle{Unsupervised Domain Adaptation With Label and
  Structural Consistency}.
\newblock \bibinfo{journal}{\emph{IEEE Transactions on Image Processing}}
  \bibinfo{volume}{25}, \bibinfo{number}{12} (\bibinfo{year}{2016}),
  \bibinfo{pages}{5552--5562}.
\newblock


\bibitem[\protect\citeauthoryear{Ionescu, Lupu, Rohm, G{\^\i}nsca, and
  M{\"u}ller}{Ionescu et~al\mbox{.}}{2018}]%
        {ionescu2018datasets}
\bibfield{author}{\bibinfo{person}{Bogdan Ionescu}, \bibinfo{person}{Mihai
  Lupu}, \bibinfo{person}{Maia Rohm}, \bibinfo{person}{Alexandru~Lucian
  G{\^\i}nsca}, {and} \bibinfo{person}{Henning M{\"u}ller}.}
  \bibinfo{year}{2018}\natexlab{}.
\newblock \showarticletitle{Datasets column: diversity and credibility for
  social images and image retrieval}.
\newblock \bibinfo{journal}{\emph{ACM SIGMultimedia Records}}
  \bibinfo{volume}{9}, \bibinfo{number}{3} (\bibinfo{year}{2018}),
  \bibinfo{pages}{7}.
\newblock


\bibitem[\protect\citeauthoryear{Krizhevsky, Sutskever, and Hinton}{Krizhevsky
  et~al\mbox{.}}{2012}]%
        {krizhevsky2012imagenet}
\bibfield{author}{\bibinfo{person}{Alex Krizhevsky}, \bibinfo{person}{Ilya
  Sutskever}, {and} \bibinfo{person}{Geoffrey~E Hinton}.}
  \bibinfo{year}{2012}\natexlab{}.
\newblock \showarticletitle{Imagenet classification with deep convolutional
  neural networks}. In \bibinfo{booktitle}{\emph{Advances in neural information
  processing systems}}. \bibinfo{pages}{1097--1105}.
\newblock


\bibitem[\protect\citeauthoryear{Long, Wang, Ding, Pan, and Philip}{Long
  et~al\mbox{.}}{2014a}]%
        {long2014adaptation}
\bibfield{author}{\bibinfo{person}{Mingsheng Long}, \bibinfo{person}{Jianmin
  Wang}, \bibinfo{person}{Guiguang Ding}, \bibinfo{person}{Sinno~Jialin Pan},
  {and} \bibinfo{person}{S~Yu Philip}.} \bibinfo{year}{2014}\natexlab{a}.
\newblock \showarticletitle{Adaptation regularization: A general framework for
  transfer learning}.
\newblock \bibinfo{journal}{\emph{IEEE Transactions on Knowledge and Data
  Engineering}} \bibinfo{volume}{26}, \bibinfo{number}{5}
  (\bibinfo{year}{2014}), \bibinfo{pages}{1076--1089}.
\newblock


\bibitem[\protect\citeauthoryear{Long, Wang, Ding, Sun, and Yu}{Long
  et~al\mbox{.}}{2013}]%
        {long2013transfer}
\bibfield{author}{\bibinfo{person}{Mingsheng Long}, \bibinfo{person}{Jianmin
  Wang}, \bibinfo{person}{Guiguang Ding}, \bibinfo{person}{Jiaguang Sun}, {and}
  \bibinfo{person}{Philip~S Yu}.} \bibinfo{year}{2013}\natexlab{}.
\newblock \showarticletitle{Transfer feature learning with joint distribution
  adaptation}. In \bibinfo{booktitle}{\emph{Proceedings of the IEEE
  International Conference on Computer Vision}}. \bibinfo{pages}{2200--2207}.
\newblock


\bibitem[\protect\citeauthoryear{Long, Wang, Ding, Sun, and Yu}{Long
  et~al\mbox{.}}{2014b}]%
        {long2014transfer}
\bibfield{author}{\bibinfo{person}{Mingsheng Long}, \bibinfo{person}{Jianmin
  Wang}, \bibinfo{person}{Guiguang Ding}, \bibinfo{person}{Jiaguang Sun}, {and}
  \bibinfo{person}{Philip~S Yu}.} \bibinfo{year}{2014}\natexlab{b}.
\newblock \showarticletitle{Transfer joint matching for unsupervised domain
  adaptation}. In \bibinfo{booktitle}{\emph{Proceedings of the IEEE Conference
  on Computer Vision and Pattern Recognition}}. \bibinfo{pages}{1410--1417}.
\newblock


\bibitem[\protect\citeauthoryear{Long, Wang, Sun, and Philip}{Long
  et~al\mbox{.}}{2015}]%
        {long2015domain}
\bibfield{author}{\bibinfo{person}{Mingsheng Long}, \bibinfo{person}{Jianmin
  Wang}, \bibinfo{person}{Jiaguang Sun}, {and} \bibinfo{person}{S~Yu Philip}.}
  \bibinfo{year}{2015}\natexlab{}.
\newblock \showarticletitle{Domain invariant transfer kernel learning}.
\newblock \bibinfo{journal}{\emph{IEEE Transactions on Knowledge and Data
  Engineering}} \bibinfo{volume}{27}, \bibinfo{number}{6}
  (\bibinfo{year}{2015}), \bibinfo{pages}{1519--1532}.
\newblock


\bibitem[\protect\citeauthoryear{Pan, Tsang, Kwok, and Yang}{Pan
  et~al\mbox{.}}{2011}]%
        {pan2011domain}
\bibfield{author}{\bibinfo{person}{Sinno~Jialin Pan}, \bibinfo{person}{Ivor~W
  Tsang}, \bibinfo{person}{James~T Kwok}, {and} \bibinfo{person}{Qiang Yang}.}
  \bibinfo{year}{2011}\natexlab{}.
\newblock \showarticletitle{Domain adaptation via transfer component analysis}.
\newblock \bibinfo{journal}{\emph{IEEE Transactions on Neural Networks}}
  \bibinfo{volume}{22}, \bibinfo{number}{2} (\bibinfo{year}{2011}),
  \bibinfo{pages}{199--210}.
\newblock


\bibitem[\protect\citeauthoryear{Pan and Yang}{Pan and Yang}{2010}]%
        {pan2010survey}
\bibfield{author}{\bibinfo{person}{Sinno~Jialin Pan} {and}
  \bibinfo{person}{Qiang Yang}.} \bibinfo{year}{2010}\natexlab{}.
\newblock \showarticletitle{A survey on transfer learning}.
\newblock \bibinfo{journal}{\emph{Knowledge and Data Engineering, IEEE
  Transactions on}} \bibinfo{volume}{22}, \bibinfo{number}{10}
  (\bibinfo{year}{2010}), \bibinfo{pages}{1345--1359}.
\newblock


\bibitem[\protect\citeauthoryear{Quanz and Huan}{Quanz and Huan}{2009}]%
        {quanz2009large}
\bibfield{author}{\bibinfo{person}{Brian Quanz} {and} \bibinfo{person}{Jun
  Huan}.} \bibinfo{year}{2009}\natexlab{}.
\newblock \showarticletitle{Large margin transductive transfer learning}. In
  \bibinfo{booktitle}{\emph{Proceedings of the 18th ACM conference on
  Information and knowledge management}}. ACM, \bibinfo{pages}{1327--1336}.
\newblock


\bibitem[\protect\citeauthoryear{Saenko, Kulis, Fritz, and Darrell}{Saenko
  et~al\mbox{.}}{2010}]%
        {saenko2010adapting}
\bibfield{author}{\bibinfo{person}{Kate Saenko}, \bibinfo{person}{Brian Kulis},
  \bibinfo{person}{Mario Fritz}, {and} \bibinfo{person}{Trevor Darrell}.}
  \bibinfo{year}{2010}\natexlab{}.
\newblock \showarticletitle{Adapting visual category models to new domains}. In
  \bibinfo{booktitle}{\emph{European conference on computer vision}}. Springer,
  \bibinfo{pages}{213--226}.
\newblock


\bibitem[\protect\citeauthoryear{Sun, Feng, and Saenko}{Sun
  et~al\mbox{.}}{2016}]%
        {sun2016return}
\bibfield{author}{\bibinfo{person}{Baochen Sun}, \bibinfo{person}{Jiashi Feng},
  {and} \bibinfo{person}{Kate Saenko}.} \bibinfo{year}{2016}\natexlab{}.
\newblock \showarticletitle{Return of Frustratingly Easy Domain Adaptation.}.
  In \bibinfo{booktitle}{\emph{AAAI}}, Vol.~\bibinfo{volume}{6}.
  \bibinfo{pages}{8}.
\newblock


\bibitem[\protect\citeauthoryear{Sun and Saenko}{Sun and Saenko}{2015}]%
        {sun2015subspace}
\bibfield{author}{\bibinfo{person}{Baochen Sun} {and} \bibinfo{person}{Kate
  Saenko}.} \bibinfo{year}{2015}\natexlab{}.
\newblock \showarticletitle{Subspace Distribution Alignment for Unsupervised
  Domain Adaptation.}. In \bibinfo{booktitle}{\emph{BMVC}}.
  \bibinfo{pages}{24--1}.
\newblock


\bibitem[\protect\citeauthoryear{Sun and Saenko}{Sun and Saenko}{2016}]%
        {sun2016deep}
\bibfield{author}{\bibinfo{person}{Baochen Sun} {and} \bibinfo{person}{Kate
  Saenko}.} \bibinfo{year}{2016}\natexlab{}.
\newblock \showarticletitle{Deep coral: Correlation alignment for deep domain
  adaptation}. In \bibinfo{booktitle}{\emph{European Conference on Computer
  Vision}}. Springer, \bibinfo{pages}{443--450}.
\newblock


\bibitem[\protect\citeauthoryear{Tahmoresnezhad and Hashemi}{Tahmoresnezhad and
  Hashemi}{2016}]%
        {tahmoresnezhad2016visual}
\bibfield{author}{\bibinfo{person}{Jafar Tahmoresnezhad} {and}
  \bibinfo{person}{Sattar Hashemi}.} \bibinfo{year}{2016}\natexlab{}.
\newblock \showarticletitle{Visual domain adaptation via transfer feature
  learning}.
\newblock \bibinfo{journal}{\emph{Knowledge and Information Systems}}
  (\bibinfo{year}{2016}), \bibinfo{pages}{1--21}.
\newblock


\bibitem[\protect\citeauthoryear{Tzeng, Hoffman, Zhang, Saenko, and
  Darrell}{Tzeng et~al\mbox{.}}{2014}]%
        {tzeng2014deep}
\bibfield{author}{\bibinfo{person}{Eric Tzeng}, \bibinfo{person}{Judy Hoffman},
  \bibinfo{person}{Ning Zhang}, \bibinfo{person}{Kate Saenko}, {and}
  \bibinfo{person}{Trevor Darrell}.} \bibinfo{year}{2014}\natexlab{}.
\newblock \showarticletitle{Deep domain confusion: Maximizing for domain
  invariance}.
\newblock \bibinfo{journal}{\emph{arXiv preprint arXiv:1412.3474}}
  (\bibinfo{year}{2014}).
\newblock


\bibitem[\protect\citeauthoryear{Vapnik and Vapnik}{Vapnik and Vapnik}{1998}]%
        {vapnik1998statistical}
\bibfield{author}{\bibinfo{person}{Vladimir~Naumovich Vapnik} {and}
  \bibinfo{person}{Vlamimir Vapnik}.} \bibinfo{year}{1998}\natexlab{}.
\newblock \bibinfo{booktitle}{\emph{Statistical learning theory}}.
  Vol.~\bibinfo{volume}{1}.
\newblock \bibinfo{publisher}{Wiley New York}.
\newblock


\bibitem[\protect\citeauthoryear{Wang et~al\mbox{.}}{Wang
  et~al\mbox{.}}{2018}]%
        {transferlearning}
\bibfield{author}{\bibinfo{person}{Jindong Wang} {et~al\mbox{.}}}
  \bibinfo{year}{2018}\natexlab{}.
\newblock \bibinfo{title}{Everything about Transfer Learning and Domain
  Adapation}.
\newblock \bibinfo{howpublished}{\url{http://transferlearning.xyz}}.
  (\bibinfo{year}{2018}).
\newblock


\bibitem[\protect\citeauthoryear{Wang, Chen, Hao, Feng, and Shen}{Wang
  et~al\mbox{.}}{2017}]%
        {wang2017balanced}
\bibfield{author}{\bibinfo{person}{Jindong Wang}, \bibinfo{person}{Yiqiang
  Chen}, \bibinfo{person}{Shuji Hao}, \bibinfo{person}{Wenjie Feng}, {and}
  \bibinfo{person}{Zhiqi Shen}.} \bibinfo{year}{2017}\natexlab{}.
\newblock \showarticletitle{Balanced distribution adaptation for transfer
  learning}. In \bibinfo{booktitle}{\emph{Data Mining (ICDM), 2017 IEEE
  International Conference on}}. IEEE, \bibinfo{pages}{1129--1134}.
\newblock


\bibitem[\protect\citeauthoryear{Xu, Fang, Wu, Li, and Zhang}{Xu
  et~al\mbox{.}}{2016}]%
        {xu2016discriminative}
\bibfield{author}{\bibinfo{person}{Yong Xu}, \bibinfo{person}{Xiaozhao Fang},
  \bibinfo{person}{Jian Wu}, \bibinfo{person}{Xuelong Li}, {and}
  \bibinfo{person}{David Zhang}.} \bibinfo{year}{2016}\natexlab{}.
\newblock \showarticletitle{Discriminative transfer subspace learning via
  low-rank and sparse representation}.
\newblock \bibinfo{journal}{\emph{IEEE Transactions on Image Processing}}
  \bibinfo{volume}{25}, \bibinfo{number}{2} (\bibinfo{year}{2016}),
  \bibinfo{pages}{850--863}.
\newblock


\bibitem[\protect\citeauthoryear{Xu, Pan, Xiong, Wu, Luo, Min, and Song}{Xu
  et~al\mbox{.}}{2017}]%
        {xu2017unified}
\bibfield{author}{\bibinfo{person}{Yonghui Xu}, \bibinfo{person}{Sinno~Jialin
  Pan}, \bibinfo{person}{Hui Xiong}, \bibinfo{person}{Qingyao Wu},
  \bibinfo{person}{Ronghua Luo}, \bibinfo{person}{Huaqing Min}, {and}
  \bibinfo{person}{Hengjie Song}.} \bibinfo{year}{2017}\natexlab{}.
\newblock \showarticletitle{A Unified Framework for Metric Transfer Learning}.
\newblock \bibinfo{journal}{\emph{IEEE Transactions on Knowledge and Data
  Engineering}} (\bibinfo{year}{2017}).
\newblock


\bibitem[\protect\citeauthoryear{Zhang, Li, and Ogunbona}{Zhang
  et~al\mbox{.}}{2017}]%
        {zhang2017joint}
\bibfield{author}{\bibinfo{person}{Jing Zhang}, \bibinfo{person}{Wanqing Li},
  {and} \bibinfo{person}{Philip Ogunbona}.} \bibinfo{year}{2017}\natexlab{}.
\newblock \showarticletitle{Joint Geometrical and Statistical Alignment for
  Visual Domain Adaptation}. In \bibinfo{booktitle}{\emph{CVPR}}.
\newblock


\bibitem[\protect\citeauthoryear{Zhuo, Wang, Zhang, and Huang}{Zhuo
  et~al\mbox{.}}{2017}]%
        {zhuo2017deep}
\bibfield{author}{\bibinfo{person}{Junbao Zhuo}, \bibinfo{person}{Shuhui Wang},
  \bibinfo{person}{Weigang Zhang}, {and} \bibinfo{person}{Qingming Huang}.}
  \bibinfo{year}{2017}\natexlab{}.
\newblock \showarticletitle{Deep Unsupervised Convolutional Domain Adaptation}.
  In \bibinfo{booktitle}{\emph{Proceedings of the 2017 ACM on Multimedia
  Conference}}. ACM, \bibinfo{pages}{261--269}.
\newblock


\end{thebibliography}

\end{document}